\documentclass[iicol,sn-mathphys-num]{sn-jnl}

\usepackage{graphicx}%
\usepackage{multirow}%
\usepackage{amsmath,amssymb,amsfonts}%
\usepackage{amsthm}%
\usepackage{mathrsfs}%
\usepackage[title]{appendix}%
\usepackage{xcolor}%
\usepackage{textcomp}%
\usepackage{manyfoot}%
\usepackage{booktabs}%
\usepackage{algorithm}%
\usepackage{algorithmicx}%
\usepackage{algpseudocode}%
\usepackage{listings}%
\usepackage{placeins}
\usepackage{makecell}
\usepackage{colortbl}
\usepackage{silence}
\usepackage{array}
\usepackage{subfig} 
\DeclareGraphicsExtensions{.pdf,.jpg,.png}

\newcommand{\bird}[1]{\textcolor{black}{#1}}

\theoremstyle{thmstyleone}%
\theoremstyle{thmstyletwo}%

\theoremstyle{thmstylethree}%

\begin{document}

\title[Article Title]{LSKNet: A Foundation Lightweight Backbone for Remote~Sensing}


\author[1]{\fnm{Yuxuan} \sur{Li}}\email{yuxuan.li.17@ucl.ac.uk}

\author[1,3]{\fnm{Xiang} \sur{Li}}\email{xiang.li.implus@nankai.edu.cn}
\equalcont{Corresponding Authors}

\author[1]{\fnm{Yimian} \sur{Dai}}\email{yimian.dai@gmail.com}

\author[1,3]{\fnm{Qibin} \sur{Hou}}\email{andrewhoux@gmail.com}

\author[2]{\fnm{Li} \sur{Liu}}\email{li.liu@oulu.fi}

\author[2]{\fnm{Yongxiang} \sur{Liu}}\email{lyx\_bible@sina.com}

\author[1,3]{\fnm{Ming-Ming} \sur{Cheng}}\email{cmm@nankai.edu.cn} 

\author[1]{\fnm{Jian} \sur{Yang}}\email{csjyang@nankai.edu.cn}
\equalcont{Corresponding Authors}

\affil[1]{\orgdiv{VCIP, CS}, \orgname{Nankai University}, \orgaddress{ \city{Tianjin}, \country{China}}}

\affil[2]{\orgname{Academy of Advanced Technology Research of Hunan}, \orgaddress{\city{Changsha}, \country{China}}}
\affil[3]{\orgdiv{NKIARI}, \orgname{Futian}, \orgaddress{ \city{Shenzhen}, \country{China}}}


\abstract{
Remote sensing images pose distinct challenges for downstream tasks due to their inherent complexity. While a considerable amount of research has been dedicated to remote sensing classification, object detection, semantic segmentation and \bird{change detection}, most of these studies have overlooked the valuable prior knowledge embedded within remote sensing scenarios.
  Such prior knowledge can be useful because remote sensing objects 
  may be mistakenly recognized without referencing a sufficiently 
  long-range context, which can vary for different objects. 
  This paper considers these priors and proposes a lightweight  
  Large Selective Kernel Network (LSKNet) backbone. 
  LSKNet can dynamically adjust its large spatial receptive field to better
  model the ranging context of various objects in remote sensing scenarios. 
  To our knowledge, 
  large and selective kernel mechanisms have not been previously explored 
  in remote sensing images.
  Without bells and whistles, our lightweight 
  LSKNet backbone network sets new state-of-the-art scores on standard remote sensing classification, object detection, semantic segmentation \bird{and change detection} benchmarks. Our comprehensive analysis further validated the significance of the identified priors and the effectiveness of LSKNet. The code is available at \url{https://github.com/zcablii/LSKNet}.
  }

\keywords{Remote sensing, CNN backbone, Large kernel, Attention, Object detection, Semantic segmentation. }

\maketitle

\section{Introduction}\label{sec:intro}
Remote sensing images present unique challenges for downstream tasks due to their complex nature, including high resolution, random orientation, large intraclass variation, multiscale scenes, and dense small objects. To tackle these challenges, extensive research has been conducted, focusing on various approaches such as feature ensemble techniques~\cite{MBLANet,MGML,ESD,KFBNet} and large-scale pretraining~\cite{RSP,wang_advancing_2022,RinMo} for classification. Additionally, methods addressing rotation variance~\cite{han_redet_2021,yang_r3det_nodate,yang_rethinking_2021}, or employing new oriented box encoding~\cite{xie_oriented_2021,xu_gliding_2021} have been proposed for object detection tasks. Furthermore, the integration of multi-scale feature fusion~\cite{UNetFormer,danet,EaNet,FANet,ABCNet,YOLOMS,WenhuaZhang-IOD1} techniques has been explored to enhance the performance of detection and segmentation tasks. 
\bird{With the rapid development of large models like SAM~\cite{sam} and LLaVA~\cite{llava}, numerous works utilize the powerful general knowledge of these models for robust downstream task fine-tuning~\cite{rsprompter,geochat}, achieving remarkable performances.}
 
\begin{figure}[!t]
  \centering
  \includegraphics[width=0.88\linewidth]{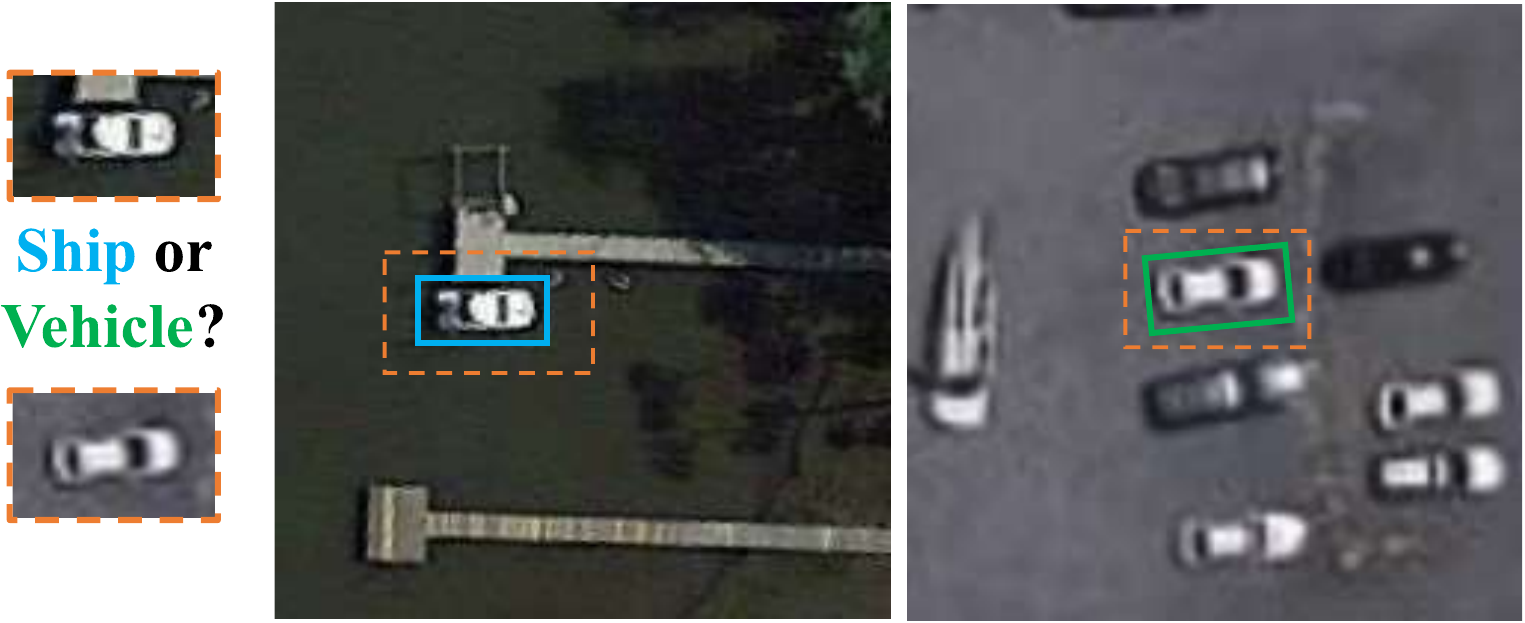}
  
  \vspace{4pt}
  \caption{Successfully detecting remote sensing objects requires using 
    a wide range of contextual information. 
    Detectors with a limited receptive field may easily lead to 
    incorrect results. 
  }\label{fig:finding_1}
\end{figure}

\begin{figure}[t]
  \centering
  \includegraphics[width=0.85\linewidth]{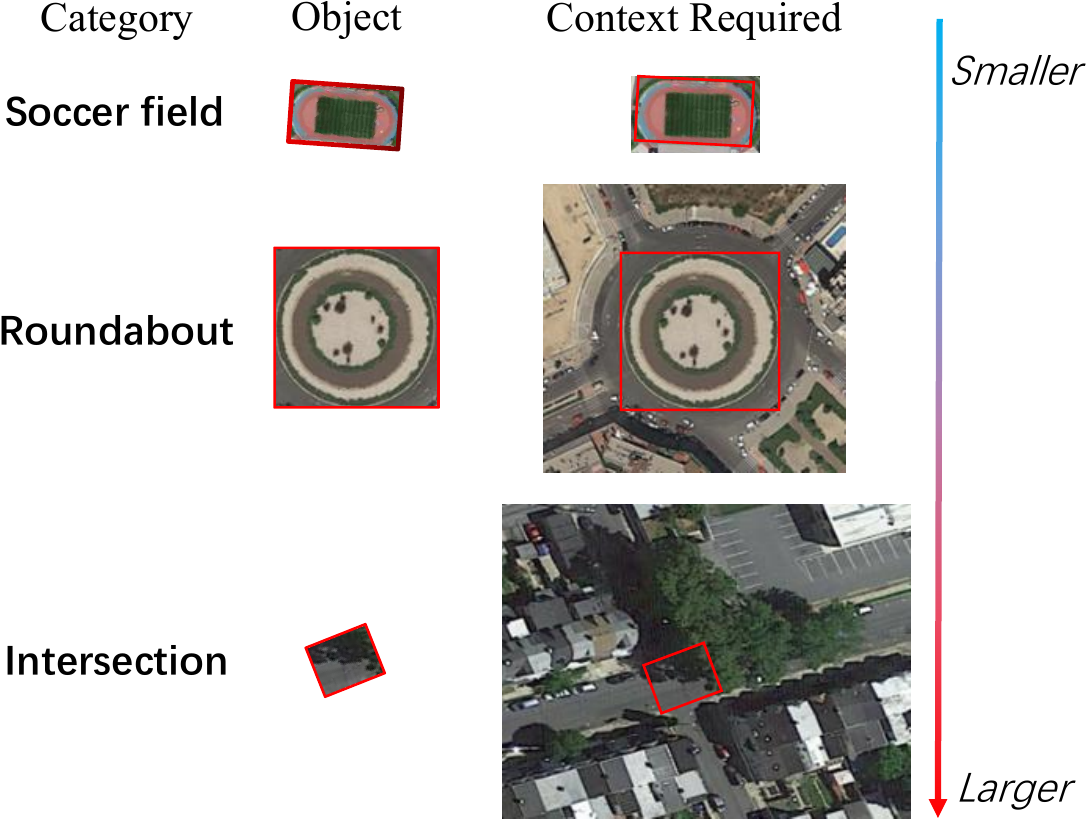}
  \vspace{4pt}
  \caption{The wide range of contextual information required for 
    different object types is very different by human criteria. 
    The objects with red boxes are the exact ground-truth annotations.
  }
  \label{fig:finding_2}
\end{figure}

Despite these advances, relatively few works have considered the
strong prior knowledge of remote sensing images to build an efficient foundation model. 
Aerial images are typically captured at high resolutions from a bird's eye view.
In particular, most objects in aerial images may be small and
difficult to identify based on their appearance alone. 
Instead, recognizing these objects relies on their context, 
as the surrounding environment can provide valuable clues about
their shape, orientation, and other characteristics. 
According to an analysis of the remote sensing data, 
we identify two important priors:
\begin{enumerate}
    \item {\bf Accurate recognition
    often requires a wide range of contextual information.} 
    As illustrated in Fig.~\ref{fig:finding_1}, 
    the limited context used by object detectors in remote sensing images 
    can often lead to incorrect classifications. 
    Rather than their appearance, 
    the context distinguishes the ship from the vehicle.
    \item {\bf The contextual information required for different objects 
    is very different.}
    As shown in Fig.~\ref{fig:finding_2}, 
    the soccer field requires relatively less contextual information 
    because of the unique distinguishable court borderlines.
    In contrast, the roundabout may require more context information 
    to distinguish between gardens and ring-like buildings.
    Intersections, especially those partially covered by trees, 
    require an extremely large receptive field due to 
    the long-range dependencies between the intersecting roads. 
\end{enumerate}

To address the challenge of accurately recognizing objects 
in remote sensing images, 
which often require a wide and dynamic range of contextual information, 
we propose a novel lightweight backbone network called 
Large Selective Kernel Network (LSKNet). 
%
Our approach employs a dynamic modulation of the receptive field within the feature extraction backbone, which allows for a more efficient accommodation and processing of the diverse, wide-ranging context that is necessitated.
This is achieved through a spatial selective mechanism, 
which efficiently weights the features processed by a sequence of 
large depth-wise kernels and then spatially merge them. 
The weights of these kernels are determined dynamically based on the input, 
allowing the model to use different large kernels adaptively 
and adjust the receptive field for each object in space as needed.

This paper presents an extended version of our previous work, \textbf{LSKNet}~\cite{lsknet}. Specifically, we have conducted further experiments to evaluate the generalization ability of our proposed LSKNet backbone across a wide range of remote sensing applications, including remote sensing scene classification on the UCM~\cite{ucm}, AID~\cite{aid}, and NWPU~\cite{nwpu} datasets, object detection on synthetic aperture radar modality dataset SAR-Aircraft~\cite{SAR_AIRcraft}, semantic segmentation tasks on the Potsdam~\cite{Potsdam}, Vaihingen~\cite{Vaihingen}, LoveDA~\cite{LoveDA}, UAVid~\cite{UAVid} and GID~\cite{GID} datasets, \bird{as well as change detection tasks on the LEVIR-CD~\cite{levir_STANet} and S2Looking~\cite{S2Looking} datasets.} Furthermore, we conducted a thorough and comprehensive comparison between LSKNet and SKNet to highlight the differences and advantages of LSKNet.

In summary, our contributions can be categorized into \textbf{FOUR} main aspects:
\begin{itemize}
    \item We have identified two significant priors present in remote sensing data.
    \item To our knowledge, the proposed LSKNet backbone is the first to explore the utilization of large and selective kernels to exactly leverage the aforementioned priors in downstream tasks of remote sensing.  
    \item Despite its simplicity and lightweight nature, our model achieves state-of-the-art performance on three prominent remote sensing tasks across 14 widely used public datasets, including remote sensing \textit{scene classification} (UCM~\cite{ucm}, AID~\cite{aid}, NWPU~\cite{nwpu}), \textit{object detection} (DOTA~\cite{dota_set}, HRSC2016~\cite{HRSC2016},
    FAIR1M~\cite{sun_fair1m_2022}, SAR-Aircraft~\cite{SAR_AIRcraft}), \textit{semantic segmentation} (Potsdam~\cite{Potsdam}, Vaihingen~\cite{Vaihingen}, LoveDA~\cite{LoveDA}, UAVid~\cite{UAVid}, GID~\cite{GID}) and \textit{change detection} (LEVIR-CD~\cite{levir_STANet}, S2Looking~\cite{S2Looking}).
    \item We provide a comprehensive analysis of our approach, further validating the importance of the identified priors and the effectiveness of the LSKNet model in addressing remote sensing challenges.
\end{itemize}

\section{Related Work}
\subsection{Remote Sensing}

\textbf{Remote Sensing Scene Classification.} Scene classification~\cite{MGML,KFBNet,RSP,wang_advancing_2022,su2023lightweight,sun2023unbiased} in remote sensing images is a challenging task due to the presence of complex backgrounds and significant intraclass variation. To address this challenge, several models, such as MGML~\cite{MGML}, ESD~\cite{ESD}, and KFBNet~\cite{KFBNet}, have been proposed. These models aim to leverage ensemble techniques that incorporate multi-level features to improve classification performance.
With the emergence of Vision Transformer (ViT)~\cite{dosovitskiy2020image}, there has been a rise in large ViT-based models~\cite{CTNet,rs13030516}. 
Moreover, recent high-performance ViT-based models such as RSP-ViTAE~\cite{RSP,vitaev2} and RVSA~\cite{wang_advancing_2022} have been pretrained on large-scale remote sensing datasets, millionAID\cite{maid}, further advancing the capabilities in this field.

However, feature ensembles often introduce multiple branches in the backbones, which is complex and computationally inefficient. Similarly, using ViT-based backbones can lead to heavy and resource-intensive, which may not be suitable for certain practical applications.

\textbf{Remote Sensing Object Detection.} Remote sensing object detection~\cite{zaidi_survey_2022,mei2023d2anet,rssurvey2023,WenhuaZhang-MOT1,WenhuaZhang-SOT1,li2024sardet} focuses on identifying and locating objects of interest in aerial images. One recent mainstream trend is to generate bounding boxes that accurately fit the orientation of the objects being detected. Consequently, a significant amount of research has focused on improving the representation of oriented bounding boxes for remote sensing object detection. Several notable detection frameworks have been introduced to mitigate the rotation variance inherent in CNN network, including the RoI transformer~\cite{ding_learning_2019}, Oriented RCNN~\cite{xie_oriented_2021}, S$^2$A network~\cite{han_align_2020}, DRN~\cite{pan_dynamic_2020} and R3Det~\cite{yang_r3det_nodate}. Oriented RCNN~\cite{xie_oriented_2021} and Gliding Vertex~\cite{xu_gliding_2021} have made significant contributions by introducing new box encoding systems 
to address the instability of training losses caused by 
rotation angle periodicity. Furthermore, techniques such as GWD~\cite{yang_rethinking_2021},KLD~\cite{yang_learning_2021} and LD \cite{23PAMI-LocDistill} have been developed to tackle the discontinuity of regression loss or enhance the localization quality of bounding boxes.

While these approaches have achieved promising results in addressing 
the issue of rotation variance, 
they do not consider the strong and valuable prior information 
presented in aerial images. 
Instead, our approach uses the large kernel 
and spatial selective mechanism to better model these priors
without modifying the current detection framework.

\newcommand{\addImg}[3]{\subfloat[#1]{\label{#2}\includegraphics[page=#3,height=.22\linewidth]{arch_comp.pdf}}}
\begin{figure*}[!htb]
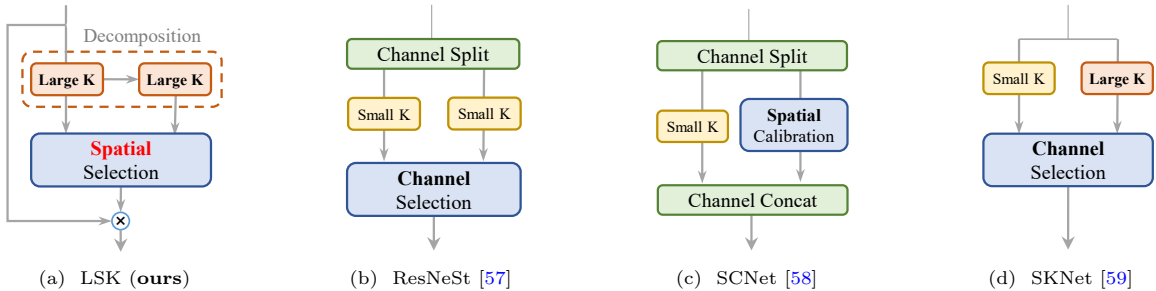

  \centering
  \tiny

  \addImg{ \footnotesize LSK (\textbf{ours})}{fig:slk_net}{4} \hfill
  \addImg{ \footnotesize ResNeSt \cite{resnest}}{fig:resnest}{3} \hfill
  \addImg{ \footnotesize SCNet \cite{liu_improving_2020}}{fig:SCNet}{2}  \hfill
  \addImg{ \footnotesize SKNet \cite{li2019selective}}{fig:Sknet}{1} \\
  \vspace{4pt}
  \caption{Architectural comparison between our proposed LSK module 
    and other selective mechanism modules. 
    K: Kernel.  
  }\label{fig:arch_compare}
\end{figure*}

\textbf{Remote Sensing Semantic Segmentation. } The most recent advancements in remote sensing semantic segmentation models have primarily focused on employing attention mechanisms and multi-scale feature fusion techniques~\cite{UNetFormer,danet,EaNet,FANet,ABCNet,BANet,MAResUNet,zhang2020causal}. These approaches effectively aggregate both fine-grained details and coarse-grained semantics, resulting in substantial improvements in segmentation performance. Consequently, it becomes evident that incorporating large receptive field semantics for multi-scale feature fusion plays a crucial role in segmentation tasks.

Despite the success achieved by existing approaches, it is often observed that they overlook the valuable \textit{prior 2)} mentioned earlier. 
In contrast, our proposed backbone model considers the valuable priors in remote sensing images, which offers more flexible multi-range receptive field features to address this limitation. 

\textbf{Remote Sensing Change Detection. } \bird{Remote sensing change detection aims to segment regions with semantic changes of interest from a pair of co-located images acquired at different times. Mainstream methods treat this task as a specialized form of segmentation with two input images. These methods involve fusing~\cite{daudt2018fully,snunet,changeformer,codegoni2023tinycd,DSIFN_IFNet} or interacting~\cite{changer,zhao2023exchanging,bit,lin2024diformer} the features of the bi-temporal images within the model's feature flow, and then using a segmentation head to generate the final change maps. } 

\bird{Numerous recent change detection frameworks~\cite{changer,wang2024mtp} demonstrate that more powerful backbones significantly improve performance, suggesting that the effectiveness and efficiency of backbone feature extraction remains a key factor in enhancing change detection models.}

\subsection{Large Kernel Networks}
Transformer-based~\cite{vaswani2017attention} models, such as the Vision Transformer (ViT) \cite{dosovitskiy2020image,wang_advancing_2022}, 
Swin transformer~\cite{liu2021swin,9736956,rs13245100,MIR-2022-07-224}, 
and pyramid transformer~\cite{pvt,wu2022p2t}, 
have gained popularity in computer vision. 
Research~\cite{Ranftl_2021_ICCV, DBLP, Zheng_2021_CVPR, luo_understanding_2016,ConcealedSurvey}
has demonstrated that the large receptive field is a key factor 
in their success.
Recent work has shown that well-designed convolutional 
networks with large receptive fields can also be highly competitive with 
transformer-based models. 
For example, ConvNeXt~\cite{liu2022convnet} uses 7$\times$7 depth-wise 
convolutions in its backbone, 
resulting in significant performance improvements in downstream tasks. 
In addition, RepLKNet~\cite{ding2022scaling} even uses a 31$\times$31 
convolutional kernel via re-parameterization, 
achieving compelling performance. 
A subsequent work SLaK~\cite{liu2022more}, 
further expands the kernel size to 51$\times$51 through 
kernel decomposition and sparse group techniques. 
RF-Next \cite{23PAMI-RF-Next} automatically searches for a fixed large kernel 
for various tasks.
VAN~\cite{guo_visual_2022} introduces an efficient decomposition of 
large kernels as convolutional attention. 
Similarly, SegNeXt~\cite{guo_segnext_2022} and 
Conv2Former~\cite{hou_conv2former_2022} demonstrate that 
large kernel convolution plays an important role in modulating 
the convolutional features with a richer context. 

Although large kernel convolutions have received attention in 
general object recognition, 
there has been a lack of research examining their significance in 
remote sensing detection. 
As previously noted in \ref{sec:intro}, 
aerial images possess unique characteristics that make large kernels 
particularly well-suited for remote sensing. 
As far as we know, our work represents the first attempt to 
introduce large kernel convolutions for remote sensing and 
to examine their importance in this field.

\subsection{Attention/Selective Mechanism}

The attention mechanism~\cite{cvm22att} is a simple but effective way 
to enhance neural representations for various tasks. 
The channel attention SE block~\cite{se_net} uses global average information 
to reweight feature channels, 
while spatial attention modules like GENet~\cite{genet}, 
GCNet~\cite{cao_gcnet_2019}, CTNet~\cite{ctnet_Zechao} and SGE~\cite{li2022spatial} 
enhance a network's ability to model context information via spatial masks. 
CBAM~\cite{ferrari_cbam_2018} and BAM~\cite{Park2018BAMBA} combine 
both channel and spatial attention. 

\bird{
Self-attention mechanisms, originally popularized in natural language processing~\cite{vaswani2017attention}, have recently gained traction in computer vision as well. Vision Transformers (ViT)~\cite{dosovitskiy2020image} leverage self-attention to capture global dependencies and contextual information across an image. In recent years, models using self-attention mechanisms achieve highly competitive performance in natural image classification~\cite{srivastava2024omnivec}, detection~\cite{carion2020end}, and segmentation~\cite{sam}. However, in many remote sensing imagery tasks, such as object detection and segmentation, global contextual information is not always necessary. For instance, to detect a car, the information about a river hundreds of meters away is not useful. Therefore, recent work has focused on incorporating locality priors into Transformer models, such as Swin~\cite{Swin}, PVT~\cite{pvt,PVT_v2}, HiViT~\cite{zhang2022hivit}, and ViTAE~\cite{Vitae}. These models offer advantages in computational efficiency and optimization compared to vanilla ViT in remote sensing scenarios~\cite{wang_advancing_2022,yu2024spatial}.}

In addition to attention mechanisms, 
kernel selection is a self-adaptive and effective technique for 
dynamic context modelling. 
CondConv~\cite{yang2019condconv} and Dynamic convolution~\cite{chen2020dynamic} 
use parallel kernels to adaptively aggregate features from 
multiple convolution kernels. 
SKNet~\cite{li2019selective} introduces multiple branches 
with different convolutional kernels and 
selectively combines them along the channel dimension. 
ResNeSt~\cite{resnest} extends the idea of SKNet by partitioning 
the input feature map into several groups. 
Similarly to the SKNet, SCNet~\cite{liu_improving_2020} uses 
branch attention to capturing richer information and spatial attention 
to improve localization ability. 
Deformable Convnets~\cite{Zhu_Deformablev2, Dai_Deformable} 
introduce a flexible kernel shape for convolution units.  

Our approach bears the most similarity to SKNet~\cite{li2019selective}.
However, there are \textbf{two key distinctions} between the two methods. 
Firstly, our proposed selective mechanism relies explicitly on 
a sequence of large kernels via decomposition, 
a departure from most existing attention-based approaches. 
Secondly, our method adaptively aggregates information across large kernels 
in the spatial dimension 
rather than the channel dimension utilized by SKNet. 
This design is more intuitive and effective for remote sensing tasks
because channel-wise selection fails to model the spatial variance 
for different targets across the image space. 
The detailed structural comparisons are listed in Fig.~\ref{fig:arch_compare}.

\begin{figure*}[t]
  \centering
  \includegraphics[width=.89\linewidth]{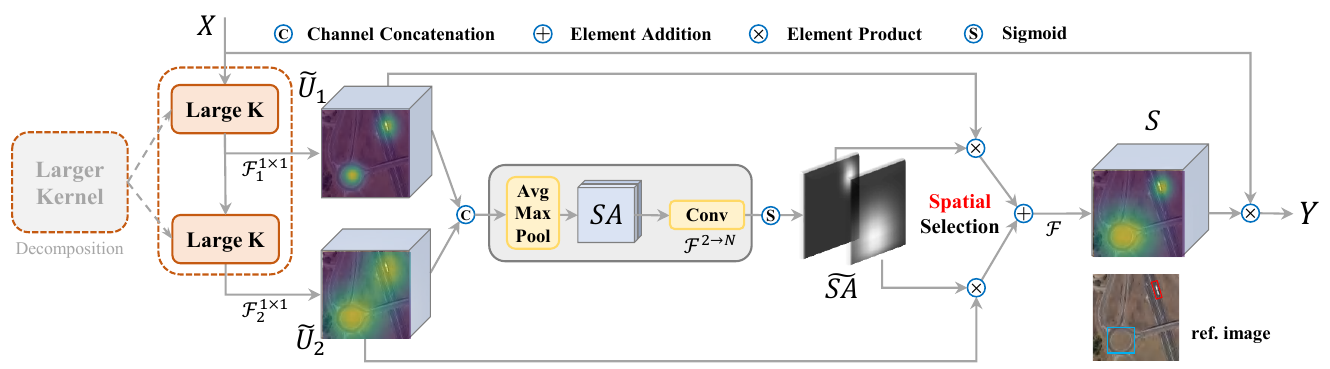}
  \vspace{4pt}
  \caption{A conceptual illustration of LSK module.}
  \label{fig:lsk_module}
\end{figure*}

\section{Methods}

\subsection{LSKNet Architecture}

 The overall architecture of the LSKNet backbone is simply built upon repeated LSK Blocks (refer to the details in Supplementary Materials). The LSK Block is inspired by ConvNeXt~\cite{liu_convnet_2s022}, MetaFormer~\cite{MetaFormer}, PVT-v2~\cite{PVT_v2}, Conv2Former~\cite{hou_conv2former_2022}, and VAN~\cite{guo_visual_2022}. Each LSK block consists of two residual sub-blocks: the Large Kernel Selection (LK Selection) sub-block and the Feed-forward Network (FFN) sub-block. 
 
 The LK Selection sub-block dynamically adjusts the network's receptive field as needed. The core LSK module (Fig.~\ref{fig:lsk_module}) is embedded in the LK Selection sub-block. It consists of a sequence of large kernel convolutions and 
a spatial kernel selection mechanism, which will be elaborated on later. The FFN sub-block is used for channel mixing and feature refinement, which consists of a sequence of a fully connected layer, a depth-wise convolution, a GELU~\cite{gelu} activation, and a second fully connected layer. 

The detailed configuration of different variants of LSKNet used in this paper 
is listed in Tab.~\ref{tab:model_variants}.
Additionally, Tab.~\ref{tab:symbol_meaning} presents a comprehensive list of important symbols, their corresponding dimensions, and their respective meanings. These symbols are extensively referenced in Fig.~\ref{fig:lsk_module} and equations presented in the subsequent sections.

\begin{table}[t]
  \centering
  \caption{\textbf{Variants of LSKNet used in this paper}. 
    $C_i$: feature channel number; 
    $D_i$: number of LSK blocks of each stage $i$.  
  }\label{tab:model_variants}
  \vspace{4pt}
  \footnotesize
  \renewcommand\arraystretch{1.2} 
  \setlength{\tabcolsep}{1mm}
  \begin{tabular}{llcr} 
    \textbf{Model} & \{{$C_1$, $C_2$, $C_3$, $C_4$}\} & \{{$D_1$, $D_2$, $D_3$, $D_4$}\} & ~~\#P \\ 
    \Xhline{1pt}
     LSKNet-T & \{32, 64, 160, 256\} & \{3, 3, 5, 2\} & ~~4.3M \\ 
     LSKNet-S & \{64, 128, 320, 512\}& \{2, 2, 4, 2\} & 14.4M \\ 
  \end{tabular} 
\end{table}

\begin{table}[t]
  \renewcommand\arraystretch{1.2} 
  \small
  \setlength{\tabcolsep}{2.5pt}
  \centering
  \caption{Symbols, dimensions, and meaning interpretations.
  }
  \label{tab:symbol_meaning}
  \vspace{4pt}
\begin{tabular}{l|c|l}
Symbol  & Dim. & Meaning                          \\
\Xhline{1pt}
$X$       & $C \times H \times W$      & input feature                    \\
$N$      & 1        & number of selection kernels      \\
$i$      & 1        & the decomposed kernel index     \\
$\widetilde{\mathbf{U}}_i$ & $C \times H \times W$      & context-rich feature             \\
$\mathbf{SA}_{max}$ & $1 \times H \times W$      & spatial attention via max pool \\
$\mathbf{SA}_{avg}$ & $1 \times H \times W$      & spatial attention via avg pool  \\
$\widetilde{\mathbf{SA}}_i$ & $N \times H \times W$      & spatial selection attentions     \\
$S$       & $C \times H \times W$     & fused attention features        \\
$Y$       &$C \times H \times W$     & output feature                  
\end{tabular}
\end{table}

\newcommand{\lrar}[1]{(#1) $\longrightarrow$ }

\begin{table}[t]
  \renewcommand\arraystretch{1.2} 
  \footnotesize
  \setlength{\tabcolsep}{2.3mm}
  \centering
  \caption{\textbf{
    Theoretical efficiency comparisons of two representative examples} 
    by expanding a single large depth-wise kernel into a sequence, 
    given channels being 64. 
    $k$: kernel size; $d$: dilation.
  }
  \label{tab:decomposite_efficiency} 
  \vspace{4pt}
  \begin{tabular}{c|l|cc} 
    RF & ($k$, $d$) sequence & \#P & FLOPs  \\ 
    \hline
    \multirow{2}{*}{{23}} & (23, 1) & ~40.4K  & ~~42.4G \\ 
    & \lrar{5,1} (7, 3)  & ~11.3K  & ~~11.9G \\ \hline
    \multirow{2}{*}{{29}} & (29, 1) & ~60.4K  & ~~63.3G  \\ 
    & \lrar{3, 1} \lrar{5, 2} (7, 3) & ~11.3K  & ~~13.6G  \\
  \end{tabular} 
\end{table}

\subsection{Large Kernel Convolutions}

According to the \textit{prior 2)} in Section \ref{sec:intro},
it is suggested to model a series of multiple long-range contexts 
for adaptive selection. 
Therefore, we propose constructing a larger kernel convolution by 
\emph{explicitly decomposing} it into a sequence of 
depth-wise convolutions with a large growing kernel and increasing dilation. 
Specifically, for the $i$-th depth-wise convolution, 
the expansion of the kernel size~$k$, dilation rate~$d$, 
and the receptive field $RF$ are defined as follows:
\begin{equation}\label{eqn:dr}
\vspace{-4pt}
  k_{i-1} \leq  k_i;~d_1 = 1,~ d_{i-1} < d_i \leq RF_{i-1} \text{,}  
\end{equation}
\begin{equation}\label{eqn:rf}
    RF_1 = k_1,~ RF_i = d_i(k_i-1) + RF_{i-1}\text{.}
\end{equation}

\noindent The increasing kernel size and dilation rate ensure that 
the receptive field expands quickly enough. 
We set an upper bound on the dilation rate to guarantee that the 
dilation convolution does not introduce gaps between feature maps. 
For instance, we can decompose a large kernel into 2 or 3 
depth-wise convolutions as in Tab.~\ref{tab:decomposite_efficiency}, 
which have a theoretical receptive field of 23 and 29, respectively.

There are two advantages of the proposed designs. First, it explicitly yields multiple features with various large receptive fields, which makes it easier for the later kernel selection. Second, sequential decomposition is more efficient than simply applying a single larger kernel. As shown in Tab.~\ref{tab:decomposite_efficiency}, under the same resulted theoretical receptive field, our decomposition greatly reduces the number of parameters compared to the standard large convolution kernels.
To obtain features with rich contextual information from different ranges for input $\mathbf{X}$, a series of decomposed depth-wise convolutions with different receptive fields are applied: 
\begin{equation}
    \mathbf{U}_0 = \mathbf{X} \text{,~~~~}  \mathbf{U}_{i+1} = \mathcal{F}^{dw}_i(\mathbf{U}_i)\text{,}
    \label{eqn:2}    
\end{equation}
where $\mathcal{F}^{dw}_i(\cdot)$ are depth-wise convolutions with kernel $k_i$ and dilation $d_i$. Assuming there are $N$ decomposed kernels, each of which is further processed by a 1$\times$1 convolution layer $\mathcal{F}^{1 \times 1}(\cdot)$:
\begin{equation}
    \widetilde{\mathbf{U}}_i =  \mathcal{F}^{1 \times 1}_i(\mathbf{U}_i), \text{ for } i \text{ in } [1,N]\text{,}
    \label{eqn:3}
\end{equation}
allowing channel mixing for each spatial feature vector.
Then, a selection mechanism is proposed
to dynamically select kernels for various objects based on the multi-scale features obtained, which would be introduced next.

\subsection{Spatial Kernel Selection}
To enhance the network's ability to focus on the most relevant spatial context regions for detecting targets, we use a spatial selection mechanism to spatially select the feature maps from large convolution kernels at different scales. 
Firstly, we concatenate the features obtained from different kernels with different ranges of receptive field:
\begin{equation}
    \widetilde{\mathbf{U}} = [\widetilde{\mathbf{U}}_1;...;\widetilde{\mathbf{U}}_i] \text{,}
    \label{eqn:4}
\end{equation}
and then efficiently extract the spatial relationship by applying channel-based average and maximum pooling (denoted as $\mathcal{P}_{avg}(\cdot)$ and $\mathcal{P}_{max}(\cdot)$) to $\widetilde{\mathbf{U}}$: %
\begin{equation}
    \mathbf{SA}_{avg} = \mathcal{P}_{avg}(\widetilde{\mathbf{U}})\text{, \ }
    \mathbf{SA}_{max} = \mathcal{P}_{max}(\widetilde{\mathbf{U}})\text{,}
    \label{eqn:5_2}
\end{equation}
where $\mathbf{SA}_{avg}$ and $\mathbf{SA}_{max}$ are the average and maximum pooled spatial feature descriptors. 
To allow information interaction among different spatial descriptors, we concatenate the spatially pooled features and use a convolution layer $\mathcal{F}^{2 \rightarrow N}(\cdot)$ to transform the pooled features (with 2 channels) into $N$ spatial attention maps:
\begin{equation}
    \widehat{\mathbf{SA}} = \mathcal{F}^{2 \rightarrow N}([\mathbf{SA}_{avg};\mathbf{SA}_{max}])\text{.}
    \label{eqn:6.2}
\end{equation}

\begin{table*}[!ht]
\renewcommand\arraystretch{1.2} 
\footnotesize 
\centering
\caption{Results of different methods for scene classification. }
\label{tab:rs_cls} 
\vspace{4pt}
\setlength{\tabcolsep}{8pt}
\begin{tabular}{l|rr|cccccc} 
Model  & \textbf{\#P $\downarrow$} & \textbf{FLOPs $\downarrow$} & \textbf{\footnotesize UCM-82} & \textbf{\footnotesize AID-28} & \textbf{\footnotesize AID-55} & \textbf{\footnotesize NWPU-19}  & \textbf{\footnotesize NWPU-28}  \\ 
\Xhline{1pt}
MSANet~\cite{MSANet}      & \begin{tiny}$>$\end{tiny}42.3M &\begin{tiny}$>$\end{tiny}164.3~~~~& 98.96  &  93.53  & 96.01  & 90.38  & 93.52 \\
ViT-B~\cite{dosovitskiy2020image}    & ~86.0M &118.9G~~~~& 99.28  &  93.81  &  96.08 &  90.96  &  93.96 \\
SCCov~\cite{SCCov}      & 13.0M &-~~~~& 99.05  & 93.12  &  96.10  &  89.30  &  92.10 \\
MA-FE~\cite{MA_FE}         & \begin{tiny}$>$\end{tiny}25.6M &\begin{tiny}$>$\end{tiny}86.3G~~~~& 99.66  & -  &  95.98  &  -  &  93.21 \\
MG-CAP~\cite{MG-CAP}     & \begin{tiny}$>$\end{tiny}42.3M &\begin{tiny}$>$\end{tiny}164.3G~~~~& 99.00   &  93.34  &  96.12 & 90.83 & 92.95 \\
LSENet~\cite{LSENet}    & ~25.9M  &\begin{tiny}$>$\end{tiny}86.3G~~~~& 99.78   &  94.41   &  96.36  & 92.23  & 93.34  \\
IDCCP~\cite{IDCCP}      & ~25.6M & 86.3G~~~~& 99.05   & 94.80   &  96.95   &  91.55 &  93.76 \\
F$^2$BRBM~\cite{F2BRBM}  & ~25.6M & 86.3G~~~~& 99.58    &  96.05   & 96.97  & 92.74   & 94.87 \\
EAM~\cite{EAM}     & \begin{tiny}$>$\end{tiny}42.3M &\begin{tiny}$>$\end{tiny}164.3~~~~& 98.98    & 94.26  & 97.06 &  91.91  & 94.29 \\
MBLANet~\cite{MBLANet}   & - &-~~~~&  99.64   &  95.60  &   97.14  &   92.32  &  94.66 \\
GRMANet~\cite{GRMANet}   &  ~54.1M & 171.4G~~~~& 99.19    & 95.43  &  97.39   & 93.19   & 94.72 \\
KFBNet~\cite{KFBNet}     & - &-~~~~& {99.88} & 95.50 &  97.40  &  93.08  &  95.11 \\
CTNet~\cite{CTNet}      & - &-~~~~& -   & 96.25 & 97.70 &  93.90 &  95.40 \\
RSP-R50~\cite{RSP}    & ~25.6M & 86.3G~~~~& 99.48  & 96.81 &  97.89 &  93.93 &  95.02 \\
RSP-Swin~\cite{RSP}     & ~27.5M & \underline{37.7G}~~~~& 99.52   & 96.83 &  98.30 &  94.02 &  94.51 \\
RSP-ViTAE~\cite{RSP}   & ~19.3M &119.1G~~~~& \underline{99.90}   & 96.91 &  98.22 &  \textbf{94.41} &  95.60 \\
RVSA~\cite{wang_advancing_2022}  & 114.4M &301.3G~~~~& -  &  \underline{97.01}  & 98.50  & 93.92  & 95.66 \\
ConvNext~\cite{liu2022convnet}  & 28.0M& 93.7G~~~ &  99.81 & 95.43 & 97.40 &  94.07 & 94.76 \\
FSCNet~\cite{FSCNet}  & 28.8M &166.1G~~~~& \textbf{100}  &  95.56  & 97.51  & 93.03  & 94.76 \\ 
UPetu~\cite{UPetu}      & 87.7M&\begin{tiny}$>$\end{tiny}322.2G~~~ & 99.05 & 96.29 & 97.06 &  92.13 &  93.79 \\
MBENet~\cite{ESD}   &  ~23.9M  &108.5G~~~ &99.81  &  96.00  &  \underline{98.54}  & 92.50  &  95.58 \\
FENet~\cite{MGML}   & ~23.9M  &92.0G~~~~& {99.86}  & 96.45 &  \textbf{98.60}  & 92.91  & 95.39  \\
\hline
\rowcolor[rgb]{0.9,0.9,0.9}$\star$ LSKNet-T  & ~~\textbf{4.3M} &\textbf{19.2G}~~~~& 99.81   & 96.80  & 98.14  & 94.07 & \underline{95.75}  \\
\rowcolor[rgb]{0.9,0.9,0.9}$\star$ LSKNet-S   & ~\underline{14.4M} &54.4G~~~~&  99.81  & \textbf{97.05}  & 98.22  & \underline{94.27}  & \textbf{95.83}  \\
\end{tabular} 
\end{table*}

\noindent 
For each of the spatial attention maps, $\widehat{\mathbf{SA}}_i$, a sigmoid activation function is applied to obtain the individual spatial selection mask for each of the decomposed large kernels:
\begin{equation}
    \widetilde{\mathbf{SA}}_i = \sigma(\widehat{\mathbf{SA}}_i)\text{,}
    \label{eqn:7}
\end{equation}
\noindent where $\sigma(\cdot)$ denotes the sigmoid function. The feature maps from the sequence of decomposed large kernels are weighted by their corresponding spatial selection masks and then fused by a convolution layer $\mathcal{F}(\cdot)$ to obtain the attention feature $\mathbf{S}$:
\begin{equation}
    \mathbf{S} = \mathcal{F}(\sum_{i=1}^N {(\widetilde{\mathbf{SA}}_i \cdot \widetilde{\mathbf{U}}_i)})\text{.}
    \label{eqn:8}
\end{equation}

\noindent The final output of the LSK module is the element-wise product between the input feature $\mathbf{X}$ and $\mathbf{S}$, similarly in~\cite{guo_visual_2022,guo_segnext_2022,hou_conv2former_2022}:
\begin{equation}
    \mathbf{Y} = \mathbf{X} \cdot \mathbf{S}\text{.}
    \label{eqn:9}
\end{equation}

\noindent Fig.~\ref{fig:lsk_module} shows a detailed conceptual illustration
of an LSK module where we intuitively demonstrate how the large selective kernel works by adaptively collecting the corresponding large receptive field for different objects.

\section{Experiments}
In this section, we report the experimental performance of our proposed model on remote sensing scene classification, object detection and semantic segmentation on a total of 11 datasets. 
In the main results, we adopt a 300-epoch backbone pretraining strategy on the Imagenet-1K~\cite{imagenet} to pursue higher accuracy, similarly to~\cite{xie_oriented_2021, han_align_2020, yang_r3det_nodate}. However, for scene classification, \bird{we follow the pretraining settings outlined in~\cite{RSP}, conducting 300 epochs of pretraining on the millionAID dataset~\cite{maid}. We directly use the official/default training, validation, and testing set splits and adhere to the mainstream settings for each benchmark to ensure fairness.} In the ablation study, we instead adopt a 100-epoch backbone pretraining strategy on the Imagenet-1K for experimental efficiency. The best score is indicated in \textbf{bold}, while the second-best score is underlined. ``FLOPs'' in this section are calculated by passing an image of 1024$\times$1024 pixels to a network. More details on experimental implementations (e.g. training schedule and data preprocessing) and result visualizations are available in Supplementary Materials.

\subsection{Scene Classification}

\subsubsection{Classification Datasets}

Mainstream of remote sensing classification research~\cite{MBLANet,F2BRBM,GRMANet,RSP} conducts experiments on the three standard scene recognition datasets including the UC Merced Land Use (UCM)~\cite{ucm} dataset, the Aerial Image Dataset (AID)~\cite{aid}, and the Image Scene Classification collected by Northwestern Polytechnical University (NWPU)~\cite{nwpu}. 

UCM is a relatively small dataset which contains only 2,100 images and 21 categories, each category has 100 images. All images are in size of 256 × 256.  

AID contains 10,000 images of 30 categories, all images are in size of 600 × 600. 

NWPU is a relatively large dataset which contains 31,500 images and 45 categories, each category has 700 images. All images are in size of 256 × 256. 

Following the mainstream of remote sensing classification works~\cite{MBLANet,F2BRBM,GRMANet,RSP}, we conduct experiments on five standard benchmarks, i.e. UCM-82, AID-28, AID-55, NWPU-19, and NWPU-28. 

\subsubsection{Classification Results}
The classification results of the compared methods are presented in Tab.~\ref{tab:rs_cls}. We compare our proposed LSKNets with 22 other state-of-the-art methods for remote sensing scene classification. 
Without any tricks such as feature ensembles in MBENet~\cite{ESD} and FENet~\cite{MGML}, our vanilla lightweight models, LSKNet-T and LSKNet-S, deliver competitive performance across multiple datasets. These results exhibit promising performance, showcasing their effectiveness for accurate scene classification across diverse scenarios and the potential for feature extraction as a backbone.

\begin{table*}[h]
\renewcommand\arraystretch{1.3} 
\tiny      
\centering
\caption{Comparison with SOTA methods on the \textbf{DOTA-v1.0} dataset with multi-scale training and testing. *: With EMA finetune similarly to the compared methods~\cite{lyu_rtmdet_2022}.}
\label{tab:dota10}
  \vspace{4pt}
\setlength{\tabcolsep}{1.2pt}
\begin{tabular}{l|c|c|c|c|ccccccccccccccc} 
Method     &\textbf{Pre.}    & \textbf{mAP$\uparrow$} & \textbf{\#P$\downarrow$} & \textbf{FLOPs$\downarrow$} & PL    & BD    & BR    & GTF   & SV    & LV    & SH    & TC    & BC    & ST    & SBF   & RA    & HA    & SP    & HC     \\ 
\Xhline{1pt}
\multicolumn{20}{l}{\textit{One-stage}}  \\ \hline
R3Det~\cite{yang_r3det_nodate}    & IN    & 76.47 & 41.9M & 336G & 89.80 & 83.77 & 48.11 & 66.77 & 78.76 & 83.27 & 87.84 & 90.82 & 85.38 & 85.51 & 65.57 & 62.68 & 67.53 & 78.56 & 72.62  \\
CFA~\cite{guo_beyond_2021}   & IN   & 76.67  & - & - & 89.08 & 83.20 & 54.37 & 66.87 & 81.23 & 80.96 & 87.17 & 90.21 & 84.32 & 86.09 & 52.34 & 69.94 & 75.52 & 80.76 & 67.96  \\
DAFNe~\cite{DAFNe}   & IN   & 76.95 &-&-  & 89.40 & \underline{86.27} & 53.70 & 60.51 & \underline{82.04} & 81.17 & 88.66 & 90.37 & 83.81 & 87.27 & 53.93 & 69.38 & 75.61 & 81.26 & 70.86  \\
SASM~\cite{SASM}   & IN   & 79.17 & - & - & 89.54 & 85.94 & 57.73 & 78.41 & 79.78 & 84.19 & 89.25 & 90.87 & 58.80 & 87.27 & 63.82 & 67.81 & 78.67 & 79.35 & 69.37  \\
AO2-DETR~\cite{dai_ao2-detr_2022} & IN   & 79.22 & 74.3M & 304G & 89.95 & 84.52 & 56.90 & 74.83 & 80.86 & 83.47 & 88.47 & 90.87 & 86.12 & \textbf{88.55} & 63.21 & 65.09 & 79.09 & \textbf{82.88} & 73.46  \\
S$^2$ANet~\cite{han_align_2020}     & IN  & 79.42  & - & - & 88.89 & 83.60 & 57.74 & 81.95 & 79.94 & 83.19 & \textbf{89.11} & 90.78 & 84.87 & 87.81 & 70.30 & 68.25 & 78.30 & 77.01 & 69.58  \\
R3Det-GWD~\cite{yang_rethinking_2021}   & IN    & 80.23 & 41.9M & 336G & 89.66 & 84.99 & 59.26 & 82.19 & 78.97 & 84.83 & 87.70 & 90.21 & 86.54 & 86.85 & \textbf{73.47} & 67.77 & 76.92 & 79.22 & 74.92  \\
RTMDet-R~\cite{lyu_rtmdet_2022}   & IN   & 80.54   & 52.3M & 205G & 88.36  & 84.96  & 57.33  & 80.46  & 80.58  & 84.88  & 88.08  & \textbf{90.90}  & 86.32  & 87.57     &  69.29     & 70.61  & 78.63  & 80.97  &  \textbf{79.24}\\
R3Det-KLD~\cite{yang_learning_2021}  & IN   & 80.63 & 41.9M & 336G & 89.92 & 85.13 & 59.19 & 81.33 & 78.82 & 84.38 & 87.50 & 89.80 & 87.33 & 87.00 & 72.57 & 71.35 & 77.12 & 79.34 & \underline{78.68}  \\
RTMDet-R~\cite{lyu_rtmdet_2022}   & CO    & 81.33  & 52.3M & 205G & 88.01 & 86.17 & 58.54 & 82.44 & 81.30 & 84.82 & 88.71 & {90.89} & \textbf{88.77} & 87.37 & 71.96 & 71.18 & 81.23 & 81.40 & 77.13  \\ 
\hline
\textit{Two-stage}                       &&      &   &         &       &       &       &       &       &       &       &       &       &       &       &       &       &       &        \\ 
\hline
SCRDet~\cite{yang_scrdet_2019}    & IN  & 72.61  & - & - & \underline{89.98} & 80.65 & 52.09 & 68.36 & 68.36 & 60.32 & 72.41 & 90.85 & 87.94 & 86.86 & 65.02 & 66.68 & 66.25 & 68.24 & 65.21  \\
ViTDet~\cite{li2022exploring} & IN & 74.41 & 103.2M & 502G  &   88.38  & 75.86 & 52.24 &  74.42  & 78.52 & 83.22 &  88.47 &  90.86 &  77.18 & 86.98 & 48.95 &  62.77 &  76.66 &  72.97  & 57.48   \\ 
Rol Trans.~\cite{ding_learning_2019}  & IN   & 74.61 & 55.1M &  200G & 88.65 & 82.60 & 52.53 & 70.87 & 77.93 & 76.67 & 86.87 & 90.71 & 83.83 & 82.51 & 53.95 & 67.61 & 74.67 & 68.75 & 61.03  \\
G.V.~\cite{xu_gliding_2021}    & IN  & 75.02   &  41.1M & 198G  & 89.64 & 85.00 & 52.26 & 77.34 & 73.01 & 73.14 & 86.82 & 90.74 & 79.02 & 86.81 & 59.55 & 70.91 & 72.94 & 70.86 & 57.32  \\
CenterMap~\cite{wang_learning_2021} & IN   & 76.03   & 41.1M & 198G  & 89.83 & 84.41 & 54.60 & 70.25 & 77.66 & 78.32 & 87.19 & 90.66 & 84.89 & 85.27 & 56.46 & 69.23 & 74.13 & 71.56 & 66.06  \\
CSL~\cite{yang_arbitrary-oriented_2020}   & IN    & 76.17   & 37.4M & 236G  & \textbf{90.25} & 85.53 & 54.64 & 75.31 & 70.44 & 73.51 & 77.62 & 90.84 & 86.15 & 86.69 & 69.60 & 68.04 & 73.83 & 71.10 & 68.93  \\
ReDet~\cite{han_redet_2021}   & IN   & 80.10  & - &  -  & 88.81 & 82.48 & 60.83 & 80.82 & 78.34 & 86.06 & 88.31 & 90.87 & \textbf{88.77} & 87.03 & 68.65 & 66.90 & 79.26 & 79.71 & 74.67  \\
DODet~\cite{DODet}   & IN  & 80.62   &  -&  - & 89.96 & 85.52 & 58.01 & 81.22 & 78.71 & 85.46 & 88.59 & {90.89} & 87.12 & 87.80 & 70.50 & 71.54 & 82.06 & 77.43 & 74.47  \\
AOPG~\cite{cheng_anchor-free_2022} & IN   & 80.66 &-& - & 89.88 & 85.57 & 60.90 & 81.51 & 78.70 & 85.29 & \underline{88.85} & {90.89} & 87.60 & 87.65 & 71.66 & 68.69 & 82.31 & 77.32 & 73.10  \\
O-RCNN~\cite{xie_oriented_2021}    & IN   & 80.87   & 41.1M &  199G & 89.84 & 85.43 & 61.09 & 79.82 & 79.71 & 85.35 & 88.82 & 90.88 & 86.68 & 87.73 & 72.21 & 70.80 & \underline{82.42} & 78.18 & 74.11  \\
KFloU~\cite{yang_kfiou_2022}  & IN   & 80.93   & 58.8M & 206G  & 89.44 & 84.41 & \underline{62.22} & 82.51 & 80.10 & \underline{86.07} & 88.68 & \textbf{90.90} & 87.32 & \underline{88.38} & \underline{72.80}  & \underline{71.95} & 78.96 & 74.95 & 75.27  \\ 
RVSA~\cite{wang_advancing_2022}  & MA & 81.24   & 114.4M & 414G & 88.97 & 85.76 & 61.46 & 81.27 & 79.98 & 85.31 & 88.30  & 90.84 & 85.06 & 87.50 & 66.77 & \textbf{73.11} & \textbf{84.75} & 81.88 & 77.58  \\ 
\hline
\rowcolor[rgb]{0.9,0.9,0.9}$\star$ LSKNet-T  & IN &    \underline{81.37}    & \textbf{21.0M} &  \textbf{124G}   &   89.14    &  84.90     & 61.78      & \underline{83.50}      & 81.54      & 85.87      & 88.64      & {90.89}      & 88.02      & 87.31      & 71.55      & 70.74      & 78.66      & 79.81      & 78.16       \\
\rowcolor[rgb]{0.9,0.9,0.9}$\star$ LSKNet-S  & IN &   \underline{81.64}   & \underline{31.0M}  & \underline{161G}  &  89.57   &  \textbf{86.34} &   \textbf{63.13} & \textbf{83.67}  & \textbf{82.20}  &  \textbf{86.10} & 88.66 & {90.89}  & 88.41  & 87.42 & 71.72 & 69.58 & 78.88  & \underline{81.77} & 76.52  \\ 
\rowcolor[rgb]{0.9,0.9,0.9}$\star$ LSKNet-S*  & IN &   \textbf{81.85}   & \underline{31.0M}  & \underline{161G}  &  89.69   &  {85.70} &   {61.47} & {83.23}  & {81.37}  &  {86.05} & 88.64 & {90.88}  & 88.49  & 87.40 & 71.67 & 71.35 & 79.19  & \underline{81.77} & \textbf{80.86}  \\ 
\end{tabular}
\end{table*}

\begin{table*}[t]
\renewcommand\arraystretch{1.2} 
\footnotesize 
\centering 
\caption{Comparison with SOTA methods on the \textbf{FAIR1M-v1.0} dataset. *: Results are referenced from FAIR1M paper~\cite{sun_fair1m_2022}.}
\label{tab:fair1m}
  \vspace{4pt}
\setlength{\tabcolsep}{1.5pt}
\begin{tabular}{c|c|c|c|c|c|c|c|c} 
Model  & G. V.*~\cite{xu_gliding_2021} & RetinaNet*~\cite{retinanet} & C-RCNN*~\cite{cascade_rcnn} & F-RCNN*~\cite{frcnn}  & RoI Trans.*~\cite{ding_learning_2019}  & O-RCNN~\cite{xie_oriented_2021}  & {\cellcolor[rgb]{0.9,0.9,0.9}} LSKNet-T & {\cellcolor[rgb]{0.9,0.9,0.9}} LSKNet-S \\ 
\hline
\textbf{mAP(\%)}     & 29.92    & 30.67    & 31.18  & 32.12  & 35.29  & 45.60   & {\cellcolor[rgb]{0.9,0.9,0.9}} \underline{46.93}    & {\cellcolor[rgb]{0.9,0.9,0.9}} \textbf{47.87}     \\
\end{tabular}
\end{table*}

\subsection{Oriented Object Detection and SAR Object Detection}
\subsubsection{Object Detection Datasets}
To evaluate the applicability of our proposed model for remote sensing detection tasks, we conducted experiments on 4 demanding datasets. These included 3 well-established oriented object detection datasets: HRSC2016~\cite{HRSC2016}, DOTA-v1.0~\cite{dota_set}, and FAIR1M-v1.0~\cite{sun_fair1m_2022}, and a highly intricate and challenging synthetic aperture radar (SAR) dataset, SAR-Aircraft~\cite{SAR_AIRcraft}.

DOTA-v1.0~\cite{dota_set} consists of 2,806 remote sensing images. It contains 188,282 instances of 15 categories: Plane (PL), Baseball diamond (BD), Bridge (BR), Ground track field (GTF), Small vehicle (SV), Large vehicle (LV), Ship (SH), Tennis court (TC), Basketball court (BC), Storage tank (ST), Soccer-ball field (SBF), Roundabout (RA), Harbor (HA), Swimming pool (SP), and Helicopter (HC). 

HRSC2016~\cite{HRSC2016} is a high-resolution remote sensing dataset which is collected for ship detection. It consists of 1,061 images which contain 2,976 instances of ships.

\begin{table}[t]
\renewcommand\arraystretch{1.2} 
\centering
\caption{Comparison with SOTA methods on \textbf{HRSC2016}. mAP (07/12): VOC 2007~\cite{voc2007}/2012~\cite{voc2012} metrics.}
\label{tab:hrsc}
  \vspace{4pt}
\footnotesize
\setlength{\tabcolsep}{0.5mm}{
\begin{tabular}{lccccc} 
Method & \footnotesize Pre. & \footnotesize\textbf{mAP(07)} & \footnotesize\textbf{mAP(12)}  & \footnotesize\textbf{\#P} & \footnotesize\textbf{FLOPs}\\ 
\Xhline{1pt}
DRN~\cite{pan_dynamic_2020}                  & IN       & -                         & 92.70       &  - &  -                \\
CenterMap~\cite{wang_learning_2021}          & IN       & -                       & 92.80      &  41.1M & 198G       \\
Rol Trans.~\cite{ding_learning_2019}        & IN    & 86.20                        & -        & 55.1M  & 200G    \\
G. V.~\cite{xu_gliding_2021}        & IN       & 88.20                        & -            & 41.1M  &  198G    \\
R3Det~\cite{yang_r3det_nodate}                 & IN       & 89.26                       & 96.01      &  41.9M  &  336G   \\
DAL~\cite{ming_dynamic_2021}                    & IN     & 89.77                       & -       &  36.4M & 216G    \\
GWD~\cite{yang_rethinking_2021}                    & IN      & 89.85                       & 97.37       &  47.4M & 456G  \\
S$^2$ANet~\cite{han_align_2020}              & IN     & 90.17                       & 95.01          & 38.6M  & 198G  \\
AOPG~\cite{cheng_anchor-free_2022}                   & IN      & 90.34                      & 96.22        &  - &  -  \\
ReDet~\cite{han_redet_2021} & IN   & 90.46 & 97.63   &  31.6M  & -     \\
O-RCNN~\cite{xie_oriented_2021}         & IN    & 90.50                        & 97.60               &  41.1M  & 199G  \\
RTMDet~\cite{lyu_rtmdet_2022}                 & CO     & \underline{90.60}                       & 97.10        & 52.3M  & 205G  \\
\hline

\rowcolor[rgb]{0.9,0.9,0.9}$\star$ LSKNet-T    & IN     &   90.54                            &        \underline{98.13}         & \textbf{21.0M}  & \textbf{124G}   \\
\rowcolor[rgb]{0.9,0.9,0.9}$\star$ LSKNet-S     & IN     &      \textbf{90.65}                            &        \textbf{98.46}         & \underline{31.0M}  & \underline{161G}   \\
\end{tabular}}
\end{table}

\begin{table}[t]
\renewcommand\arraystretch{1.2} 
\footnotesize
\setlength{\tabcolsep}{1.3mm}
\centering
\caption{The mAP results on the test set for the \textbf{SAR-Aircraft} dataset. }
\label{tab:sar}
  \vspace{4pt}
\begin{tabular}{l|ccc} 
  \textbf{RetinaNet~\cite{retinanet} 2x }& \textbf{\#P} & \textbf{mAP$_{50}$}   & \textbf{mAP$_{75}$}   \\
\Xhline{1pt}              
  ResNet-50~\cite{he2016deep}  &  25.6M & 0.469 & 0.324 \\
   PVT-Tiny~\cite{pvt}  &  13.2M & 0.498  & 0.335   \\ 
   Res2Net-50~\cite{res2net}  & 25.7M & 0.528  & 0.339  \\ 
  Swin-T~\cite{Swin}    & 28.3M & 0.586  & 0.346  \\
  ConvNeXt V2-N~\cite{Woo2023ConvNeXtVC}  & 15.0M   & 0.589 & 0.350    \\ 
   VAN-B1~\cite{guo_visual_2022}   & 13.4M  &  {0.603}   &  {0.375} \\ 
   \hline
 \rowcolor[rgb]{0.9,0.9,0.9}$\star$  LSKNet-T   & \textbf{4.3M}   & 0.582 &  0.354  \\
 \rowcolor[rgb]{0.9,0.9,0.9}$\star$  LSKNet-S    & 14.4M   & \textbf{0.624} & \textbf{0.387} \\
\multicolumn{4}{c}{ }  \\
 
  \textbf{Cascade Mask RCNN~\cite{cascade} 2x} & \textbf{\#P} &\textbf{ mAP$_{50}$}   & \textbf{mAP$_{75}$}   \\
\Xhline{1pt}
 ResNet-50~\cite{he2016deep}  &  25.6M    & 0.483 & 0.339 \\
  PVT-Tiny~\cite{pvt}    &  13.2M  & 0.502  & 0.344 \\
  Res2Net-50~\cite{res2net}  &  25.7M   & 0.544 & 0.372  \\
  ConvNeXt V2-N~\cite{Woo2023ConvNeXtVC}  & 15.0M   & 0.581  & 0.428    \\ 
  Swin-T~\cite{Swin}    & 28.3M & 0.596  & 0.416  \\
   VAN-B1~\cite{guo_visual_2022}    &  13.4M  & {0.604} & {0.457}  \\
   \hline
\rowcolor[rgb]{0.9,0.9,0.9} $\star$ LSKNet-T   & \textbf{4.3M}  & {0.586} & 0.435 \\
 \rowcolor[rgb]{0.9,0.9,0.9}$\star$ LSKNet-S  & 14.4M  & \textbf{0.614} & \textbf{0.458}
\end{tabular}
\end{table}

FAIR1M-v1.0~\cite{sun_fair1m_2022} is a recently published remote sensing dataset that consists of 15,266 high-resolution images and more than 1 million instances. It contains 5 categories and 37 sub-categories objects.

The SAR-Aircraft dataset~\cite{SAR_AIRcraft} is a recently published remote sensing dataset specifically collected for the SAR modality object detection. Different from the above 3 datasets which are in RGB modality, SAR datasets are in grayscale. It encompasses 7 distinct categories, namely A220, A320/321, A330, ARJ21, Boeing737, Boeing787, and other. The dataset comprises a training set of 3,489 images and a test set of 879 images, totalling 16,463 instances of aircraft.

\subsubsection{Detection Results} 

In our oriented object detection experiments, LSKNets are defaulting to be built within the Oriented RCNN~\cite{xie_oriented_2021} framework due to its compelling performance and efficiency.

\textbf{Results on DOTA-v1.0.} 
We compare our LSKNet with 20 state-of-the-art methods on the DOTA-v1.0 dataset, as reported in Tab.~\ref{tab:dota10}. Our LSKNet-T, LSKNet-S and LSKNet-S* achieve state-of-the-art with mAP of \textbf{81.37\%}, \textbf{81.64\%} and \textbf{81.85\%} respectively. Notably, our high-performing LSKNet-S reaches an inference speed of \textbf{18.1} FPS on 1024x1024 images with a single RTX3090 GPU.

\textbf{Results on HRSC2016.} 
We evaluated the performance of our LSKNet against 12 state-of-the-art methods on the HRSC2016 dataset. The results presented in Tab.~\ref{tab:hrsc} demonstrate that our LSKNet-S outperforms all other methods with an mAP of \textbf{90.65\%} and \textbf{98.46\%} under the PASCAL VOC 2007~\cite{voc2007} and VOC 2012~\cite{voc2012} metrics, respectively. 

\textbf{Results on FAIR1M-v1.0.} 
We compare our LSKNet against 6 other models on the FAIR1M-v1.0 dataset, as shown in Tab.~\ref{tab:fair1m}. The results reveal that our LSKNet-T and LSKNet-S perform exceptionally well, achieving state-of-the-art mAP scores of \textbf{46.93\%} and \textbf{47.87\%} respectively, surpassing all other models by a significant margin. Fine-grained category results can be found in Supplementary Materials.

\textbf{Results on SAR-Aircraft.} 
We evaluate the performance of our proposed LSKNets compared to 5 state-of-the-art backbone networks under Cascade Mask RCNN~\cite{cascade} and RetinaNet~\cite{retinanet} detection frameworks. The results are shown in Tab.~\ref{tab:sar}, which clearly shows that our proposed LSKNets provide a significant and substantial improvement in performance for SAR object detection. 

\textbf{Quantitative Analysis.}
\bird{The ViTDet, which uses the vanilla ViT backbone, has the largest computational complexity (4.0x FLOPs compared to LSKNet-T) and the second largest model size (4.9x parameters compared to LSKNet-T) among the compared models, and it performs poorly on object detection on DOTA-v1.0 dataset. Another variant of the ViT-based model, RVSA, which is based on ViTAE, incorporates multi-scale and 2D locality inductive bias and is more effective at modelling image features than the vanilla ViT backbone. Despite its effectiveness, RVSA still suffers from a heavy model size (5.4x parameters compared to LSKNet-T) and high computational complexity (3.3x FLOPs compared to LSKNet-T). Neither of these ViT-based models can outperform the lightweight LSKNet-T.
\\
LSKNet's advantages are also observed in easily mixed-up categories such as small vehicles (+2.49\%) and ships (+3.59\%) in the DOTA-v1.0 dataset (Tab.~\ref{tab:dota10}), as well as in categories that require large context information such as intersections (+2.08\%), roundabouts (+6.53\%), and bridges (+6.11\%) in the FAIR1M dataset (Tab.~S4 in the Supplementary Materials). These results further verify our identified \textit{Prior 1} and \textit{Prior 2}, and justify the effectiveness of the proposed foundation backbone model.}

\subsection{Semantic Segmentaion}

\subsubsection{Segmentaion Dataset}
Following the mainstream segmentation research~\cite{UNetFormer,BANet}, we assess the effectiveness of our proposed model in remote sensing segmentation by conducting evaluations on five standard datasets: Potsdam~\cite{Potsdam}, Vaihingen~\cite{Vaihingen}, LoveDA~\cite{LoveDA}, UAVid~\cite{UAVid} and GID~\cite{GID} dataset.

Potsdam~\cite{Potsdam}  is a high-resolution semantic segmentation dataset that consists of 38 high-resolution images. It is composed of 6 categories of semantics: impervious surface, building, low vegetation, tree, car, and one background category, clutter.   

Vaihingen~\cite{Vaihingen} is also a fine spatial resolution semantic segmentation dataset which consists of 33 high-resolution images. It has the same semantic categories as Potsdam.

LoveDA~\cite{LoveDA} is a multi-scale and complex remote sensing semantic segmentation dataset that contains 5,987 1024×1024 pixels images. Among these images, 2522 are allocated for training, 1,669 for validation, and 1,796 for online testing. The dataset consists of 7 categories of semantics: building, road, water, barren, forest, agriculture and background (Back.G.).  

UAVid~\cite{UAVid} is a high-resolution and complex Unmanned Aerial Vehicle (UAV) semantic segmentation dataset. It contains 200 training images, 70 validation images and 150 online testing images. The dataset is composed of 8 distinct categories: Clutter, Building, Road, Tree, Vegetation, Moving Car(Mo.Car), Static Car(St.Car), and Human. 

\bird{The GID~\cite{GID} dataset is a medium-resolution land cover segmentation dataset with a ground sampling distance (GSD) of 4m, containing 150 images of 7,200×6,800 pixels. Following~\cite{gid_split}, we selected the 15 pre-defined images from the original GID dataset and cropped all images into 256×256 pixels, resulting in 7,830 training images and 3,915 testing images. The dataset consists of six semantic categories: build-up, farmland, forest, meadow, water, and others.}

\begin{table}[t]
\renewcommand\arraystretch{1.2} 
\footnotesize 
\centering
\caption{Quantitative comparison results on the Potsdam test set. OA: Overall Accuracy}
\label{tab:Potsdam}
  \vspace{4pt}
\setlength{\tabcolsep}{7.2pt}
\begin{tabular}{l|ccc} 
Model  & \textbf{mF1~$\uparrow$} & \textbf{OA~$\uparrow$} & \textbf{mIOU~$\uparrow$}   \\ 
\Xhline{1pt}

ERFNet~\cite{ERFnet}   & 85.8   & 84.5 & 76.2   \\
DABNet~\cite{dabnet}       & 88.3   & 86.7 & 79.6   \\
BiSeNet~\cite{BiSeNet}   & 89.8   & 88.2 & 81.7  \\
EaNet~\cite{EaNet}      & 90.6   & 88.7 & 83.4\\
MARESU-Net~\cite{MAResUNet}  & 90.5   & 89.0 & 83.9 \\
DANet~\cite{danet}   & 88.9   & 89.1 & 80.3 \\
SwiftNet~\cite{SwiftNet}  & 91.0   & 89.3 & 83.8  \\
FANet~\cite{FANet}    & 91.3   & 89.8 & 84.2\\
ShelfNet~\cite{shelfnet} & 91.3   & 89.9 & 84.4  \\
ABCNet~\cite{ABCNet}   & 92.7   & 91.3 & 86.5 \\
Segmenter~\cite{Segmenter} & 89.2   & 88.7 & 80.7  \\
BANet~\cite{BANet}   & 92.5   & 91.0 & 86.3  \\
SwinUperNet~\cite{Swin} & 92.2   & 90.9 & 85.8  \\
UNetFormer~\cite{UNetFormer}   & 92.8   & 91.3 & \underline{86.8} \\
\hline
\rowcolor[rgb]{0.9,0.9,0.9}$\star$ LSKNet-T   & \underline{92.9} & \underline{91.7} & 86.7  \\
\rowcolor[rgb]{0.9,0.9,0.9}$\star$ LSKNet-S    & \textbf{93.1} & \textbf{92.0} & \textbf{87.2}  
\vspace{2pt}
\end{tabular}
\end{table}

\begin{table}[t]
\renewcommand\arraystretch{1.2} 
\footnotesize 
\centering
\caption{Quantitative comparison results on the Vaihingen test set. OA: Overall Accuracy}
\label{tab:Vaihingen}
  \vspace{4pt}
\setlength{\tabcolsep}{7.2pt}
\begin{tabular}{l|ccc} 
Model   & \textbf{mF1~$\uparrow$} & \textbf{OA~$\uparrow$} & \textbf{mIOU~$\uparrow$} \\ 
\Xhline{1pt}
PSPNet~\cite{PSPNet}  & 79.0 & 87.7 & 68.6  \\
ERFNet~\cite{ERFnet} & 78.9 & 85.8 & 69.1 \\
DANet~\cite{danet}  & 79.6 & 88.2 & 69.4  \\
DABNet~\cite{dabnet}  & 79.2 & 84.3 & 70.2  \\
Segmenter~\cite{Segmenter} & 84.1 & 88.1 & 73.6   \\
BOTNet~\cite{BoTNet} & 84.8 & 88.0 & 74.3   \\
FANet~\cite{FANet}  & 85.4 & 88.9 & 75.6   \\
BiSeNet~\cite{BiSeNet} & 84.3 & 87.1 & 75.8  \\
DeepLabV3+~\cite{deeplabv3}  & 87.4 & 89.0 & -  \\
ShelfNet~\cite{shelfnet}  & 87.5 & 89.8 & 78.3  \\
MARESU-Net~\cite{MAResUNet} & 87.7 & 90.1 & 78.6  \\
EaNet~\cite{EaNet}    & 87.7 & 89.7 & 78.7 \\
SwiftNet~\cite{SwiftNet} & 88.3 & 90.2 & 79.6  \\
ABCNet~\cite{ABCNet}  & 89.5 & 90.7 & 81.3 \\
BANet~\cite{BANet}  & 89.6 & 90.5 & 81.4 \\
UNetFormer~\cite{UNetFormer} & 90.4 & 91.0 & 82.7  \\
\hline
\rowcolor[rgb]{0.9,0.9,0.9}$\star$ LSKNet-T   & \underline{91.7} & \textbf{93.6} & \underline{84.9}  \\
\rowcolor[rgb]{0.9,0.9,0.9}$\star$ LSKNet-S     & \textbf{91.8} & \textbf{93.6}  & \textbf{85.1}  \\ 
\end{tabular}
\end{table}

\begin{table*}[t]
\renewcommand\arraystretch{1.2} 
\footnotesize 
\centering
\caption{Quantitative comparison results on the LoveDA test set.}
\label{tab:LoveDA}
  \vspace{4pt}
\setlength{\tabcolsep}{8.5pt}
\begin{tabular}{l|c|ccccccc} 
Method   & \textbf{mIoU~$\uparrow$} & Back.G. & Building & Road & Water & Barren & Forest & Agriculture \\
\Xhline{1pt}
Segmenter~\cite{Segmenter}      & 47.1 & 38.0       & 50.7     & 48.7 & 77.4    & 13.3   & 43.5   & 58.2      \\
SegFormer~\cite{SegFormer}    & 47.4 & 43.1      & 52.3     & 55.0 & 70.7    & 10.7   & 43.2   & 56.8  \\
DeepLabV3+~\cite{deeplabv3}     & 47.6 & 43.0       & 50.9     & 52.0 & 74.4    & 10.4   & 44.2   & 58.5     \\
UNet~\cite{UNet}    & 47.6 & 43.1       & 52.7     & 52.8 & 73.0    & 10.3   & 43.1   & 59.9      \\
UNet++~\cite{UNet++}    & 48.2 & 42.9  & 52.6   & 52.8 & 74.5    & 11.4   & 44.4   & 58.8     \\
SemanticFPN~\cite{SemanticFPN}     & 48.2 & 42.9       & 51.5     & 53.4 & 74.7    & 11.2   & 44.6   & 58.7      \\
FarSeg~\cite{FarSeg}    & 48.2  & 43.1       & 51.5     & 53.9 & 76.6    & 9.8    & 43.3   & 58.9       \\
PSPNet~\cite{PSPNet}     & 48.3  & 44.4       & 52.1     & 53.5 & 76.5    & 9.7    & 44.1   & 57.9       \\
FactSeg~\cite{FactSeg}    & 48.9 & 42.6       & 53.6     & 52.8 & 76.9    & 16.2   & 42.9   & 57.5       \\
TransUNet~\cite{Transunet}   & 48.9 & 43.0       & 56.1     & 53.7 & 78.0    & 9.3    & 44.9   & 56.9      \\
BANet~\cite{BANet}  & 49.6 & 43.7       & 51.5     & 51.1 & 76.9    & 16.6   & 44.9   & \underline{62.5}       \\
HRNet~\cite{HRNet}    & 49.8 & 44.6      & 55.3   & \underline{57.4} & 78.0    & 11.0   & 45.3   & 60.9      \\
SwinUperNet~\cite{Swin}     & 50.0 & 43.3       & 54.3     & 54.3 & 78.7    & 14.9   & 45.3   & 59.6       \\
DC-Swin~\cite{DC-Swin} & 50.6 & 41.3       & 54.5     & 56.2 & 78.1    & 14.5   & \textbf{47.2}   & 62.4        \\
UNetFormer~\cite{UNetFormer}      & 52.4 & 44.7       & 58.8     & 54.9 & 79.6    & 20.1   & 46.0   & \underline{62.5}    \\
Hi-ResNet~\cite{hi_resnet}     & 52.5 & \textbf{46.7}  & 58.3   & 55.9 & \underline{80.1}    & 17.0   & \underline{46.7}   & \textbf{62.7}  \\
\hline
\rowcolor[rgb]{0.9,0.9,0.9}$\star$ LSKNet-T  & \underline{53.2}  & 46.4  & \underline{59.5} & 57.1 & 79.9 & \underline{21.8} & 46.6 & 61.4 \\
\rowcolor[rgb]{0.9,0.9,0.9}$\star$ LSKNet-S   & \textbf{54.0}  & \textbf{46.7} & \textbf{59.9} & \textbf{58.3} & \textbf{80.2} & \textbf{24.6}  & 46.4 & 61.8 
\vspace{2pt}
\end{tabular}
\end{table*}

\begin{table*}[t]
\renewcommand\arraystretch{1.2} 
\footnotesize 
\centering
\caption{Quantitative comparison results on the UAVid test set. }
\label{tab:UAVid}
  \vspace{4pt}
\setlength{\tabcolsep}{8.pt}
\begin{tabular}{l|c|ccccccccc}
Method   & \textbf{mIoU~$\uparrow$}   & Clutter & Building & Road & Tree & Vegetation  & Mo.Car & St.Car & Human \\
\Xhline{1pt}
MSD~\cite{UAVid}        & 57.0 & 57.0    & 79.8     & 74.0 & 74.5 & 55.9       & 62.9       & 32.1      & 19.7  \\
CANet~\cite{CANet}      & 63.5 & 66.0    & 86.6     & 62.1 & 79.3 & \textbf{78.1}       & 47.8       & \textbf{68.3}      & 19.9  \\
DANet~\cite{danet}      & 60.6 & 64.9    & 85.9     & 77.9 & 78.3 & 61.5       & 59.6       & 47.4      & 9.1   \\
SwiftNet~\cite{SwiftNet}   & 61.1 & 64.1    & 85.3     & 61.5 & 78.3 & 76.4       & 51.1       & 62.1      & 15.7  \\
BiSeNet~\cite{BiSeNet}    & 61.5 & 64.7    & 85.7     & 61.1 & 78.3 & 77.3       & 48.6       & 63.4      & 17.5  \\
MANet~\cite{MAResUNet}      & 62.6 & 64.5    & 85.4     & 77.8 & 77.0 & 60.3       & 67.2       & 53.6      & 14.9  \\
ABCNet~\cite{ABCNet}     & 63.8 & 67.4    & 86.4     & 81.2 & 79.9 & 63.1       & 69.8       & 48.4      & 13.9  \\
Segmenter~\cite{Segmenter}  & 58.7 & 64.2    & 84.4     & 79.8 & 76.1 & 57.6       & 59.2       & 34.5      & 14.2  \\
SegFormer~\cite{SegFormer}  & 66.0 & 66.6    & 86.3     & 80.1 & 79.6 & 62.3       & 72.5       & 52.5      & 28.5  \\
BANet~\cite{BANet}      & 64.6 & 66.7    & 85.4     & 80.7 & 78.9 & 62.1       & 69.3       & 52.8      & 21.0  \\
BOTNet~\cite{BoTNet}     & 63.2 & 64.5    & 84.9     & 78.6 & 77.4 & 60.5       & 65.8       & 51.9      & 22.4  \\
CoaT~\cite{CoaT}       & 65.8 & 69.0    & \textbf{88.5}     & 80.0 & 79.3 & 62.0       & 70.0       & 59.1      & 18.9  \\
UNetFormer~\cite{UNetFormer} & 67.8 & 68.4    & 87.4     & 81.5 & 80.2 & 63.5       & 73.6       & 56.4      & 31.0  \\
\hline
\rowcolor[rgb]{0.9,0.9,0.9}$\star$ LSKNet-T & \underline{69.3} & \textbf{69.6} & \underline{87.9} & \underline{82.8} & \underline{80.6} & 64.8 & \textbf{77.3} & 60.2  & \underline{31.3}  \\
\rowcolor[rgb]{0.9,0.9,0.9}$\star$ LSKNet-S   & \textbf{70.0} & \textbf{69.6} & 84.8 & \textbf{82.9} & \textbf{80.9} & \underline{65.5} & \underline{76.8} & \underline{64.9}  & \textbf{31.8} \\
\end{tabular}
\end{table*}

\begin{table}[ht]
\renewcommand\arraystretch{1.2}  
\footnotesize 
\centering
\caption{Quantitative comparison results on the GID test set. }
\label{tab:gid}
  \vspace{4pt}
\setlength{\tabcolsep}{8.pt}
\begin{tabular}{l|ccc}
Backbones        & \textbf{mF1~$\uparrow$}  & \textbf{OA~$\uparrow$}   & \textbf{mIoU~$\uparrow$} \\
\Xhline{1pt}
ConvNext-v2-N~\cite{Woo2023ConvNeXtVC} & 75.1 & 78.9 & 62.5 \\
ResNet-50~\cite{he2016deep}     & 75.3 & 80.0   & 64.1 \\
Swin-T~\cite{Swin}        &    77.8  & 80.8     & 65.6     \\
ResNest-50~\cite{resnest}     & 79.7  &  80.3   & 67.2 \\
VAN-S~\cite{guo_visual_2022}         & 80.2 & \underline{82.1} & 68.2 \\
MSCAN-S~\cite{guo_segnext_2022}         & \underline{80.4} & 81.4 & \underline{68.4} \\
\hline
\rowcolor[rgb]{0.9,0.9,0.9}$\star$ LSKNet-T   & {79.4} & 81.5 & 67.2  \\
\rowcolor[rgb]{0.9,0.9,0.9}$\star$ LSKNet-S     & \textbf{83.2} & \textbf{82.3}  & \textbf{69.6}  \\  
\end{tabular}
\end{table}

\begin{table*}[t]
\renewcommand\arraystretch{1.2} 
\footnotesize 
\centering
\caption{Quantitative comparison results for change detection on LEVIR-CD and S2Looking datasets.}
\label{tab:change}
  \vspace{4pt}
\setlength{\tabcolsep}{4.pt}
\begin{tabular}{l|cccc|cccc}
\multicolumn{1}{c|}{\multirow{2}{*}{Method}} & \multicolumn{4}{c|}{\underline{~~~~~~~~~~~~~~~~~\textbf{LEVIR-CD~\cite{levir_STANet}}~~~~~~~~~~~~~~~~}}      & \multicolumn{4}{c}{\underline{~~~~~~~~~~~~~~~~~\textbf{S2Looking~\cite{S2Looking}}~~~~~~~~~~~~~~~~~}}       \\ 
& \textbf{Precision~$\uparrow$} & \textbf{~Recall~$\uparrow$~} & \textbf{~~F1~$\uparrow$~~ }   & \textbf{~~IoU~$\uparrow$~~ }  &\textbf{ Precision~$\uparrow$} & \textbf{~Recall~$\uparrow$~} & \textbf{~~F1~$\uparrow$~~}   & \textbf{~~IoU~$\uparrow$~~}   \\ 
\Xhline{1pt}
FC-EF~\cite{daudt2018fully}         & 86.91     & 80.17  & 83.40      & 71.53    & \underline{81.36 }    & 8.95   & 7.65          & 8.77          \\
FC-Siam-Conc~\cite{daudt2018fully}  & 91.99     & 76.77  & 83.69      & 71.96       & \textbf{83.29}     & 15.76  & 13.19         & 15.28         \\
FC-Siam-Di~\cite{daudt2018fully}    & 89.53     & 83.31  & 86.31      & 75.92        & 68.27     & 18.52  & 13.54         & 17.05         \\
STANet~\cite{levir_STANet}       & 83.81     & {91.00}     & 87.26  & 77.40        & 38.75     & 56.49  & 45.97         & 29.84         \\
DTCDSCN~\cite{DTCDSCN}      & 88.53     & 86.83  & 87.67        & 78.05      & 68.58     & 49.16  & 57.27         & 40.12         \\
HANet~\cite{HANet}         & 91.21     & 89.36  & 90.28           & 82.27      & 61.38     & 55.94  & 58.54         & 41.38         \\
CDNet~\cite{cdnet}        & 91.60      & 86.50   & 89.00             & 80.14      & 67.48     & 54.93  & 60.56         & 43.43      \\
CDMC~\cite{cdmc}          & 93.09      &88.07   & 90.51             & 82.67     & 64.88     & 58.15     &61.34        & 44.23 \\
IFNet~\cite{DSIFN_IFNet}        & 91.17     & 90.51  & 90.83         & 83.22        & 66.46     & 61.95  & 64.13         & 47.19         \\
SNUNet~\cite{snunet}       & 92.45     & 90.17  & 91.30              & 83.99         & 71.94     & 56.34  & 63.19         & 46.19         \\
BiT~\cite{bit}            & 91.97     & 88.62  & 90.26             & 82.26            & 74.80      & 55.56  & 63.76         & 46.80          \\
HCGMNet~\cite{hcgmnet}      & 92.96     & 90.61  & 91.77             & 84.79            & 72.51     & 57.06  & 63.87         & 46.91         \\
ChangeFormer~\cite{changeformer} & 92.59     & 89.68   & 91.11         & 83.67            & 72.82     & 56.13  & 63.39         & 46.41         \\
C2FNet~\cite{c2fnet}        & \underline{93.69}     & 89.47  & 91.83             & 84.89            & 74.84     & 54.14  & 62.83         & 45.80          \\
CGNet~\cite{CGNet}         & 93.15     & 90.90   & 92.01             & 85.21            & 70.18     & 59.38  & 64.33         & 47.41    \\
DiFormer~\cite{lin2024diformer} & \textbf{93.75} & 90.59 &  {92.15} & {85.44} & 72.39 & 61.19 &66.31  & 49.60 \\
Changer-MiT\_b0~\cite{changer}      & 93.61 & 90.56  & 92.06 & 85.29  & 73.01 & 62.04 & 67.08 & 50.47   \\  \hline
\rowcolor[rgb]{0.9,0.9,0.9}$\star$ LSKNet-T & 92.56 & \textbf{91.83} & \underline{92.19} & \underline{85.51} & 70.44&\textbf{64.46} & \underline{67.32} &\underline{50.74} \\
\rowcolor[rgb]{0.9,0.9,0.9}$\star$ LSKNet-S  & 93.34& \underline{91.23}& \textbf{92.27} & \textbf{85.65}&71.90 &\underline{63.64} & \textbf{67.52} &\textbf{50.96} \\
    
\end{tabular}
\end{table*}

\subsubsection{Segmentation Results}
We conducted a comprehensive comparison of our proposed models, LSKNet-T and LSKNet-S, against a multitude of recently proposed high-performance models on the 5 aforementioned datasets. For the Potsdam, Vaihingen, LoveDA and UAVid datasets, LSKNets are integrated within the UNetFormer~\cite{UNetFormer} framework due to its compelling performance and open-source availability. For the GID dataset, we compare various backbone models using the SegFormer framework.
Specifically, we compared our models to 14 models on the Potsdam dataset (Tab.~\ref{tab:Potsdam}), 16 models on the Vaihingen dataset (Tab.~\ref{tab:Vaihingen}), 13 models on the LoveDA dataset (Tab.~\ref{tab:LoveDA}), 16 models on the UAVid dataset (Tab.~\ref{tab:UAVid}) and 6 backbone models on the GID dataset (Tab.~\ref{tab:gid}).
Notably, our LSKNet-T and LSKNet-S models displays exceptional performance, surpassing all other state-of-the-art methods across all datasets in most major metrics.

\subsection{Change Detection}

\subsubsection{Change Detection Dataset}
\bird{Following the mainstream change detection research~\cite{changer,lin2024diformer,CGNet}, we assess the effectiveness of our proposed model in remote sensing change detection tasks by conducting evaluations on two standard datasets: LEVIR-CD ~\cite{levir_STANet} and S2Looking ~\cite{S2Looking}.}

\bird{LEVIR-CD~\cite{levir_STANet} includes 637 pairs of bi-temporal images sourced from Google Earth, with each image having a size of 1024 × 1024 pixels and a ground sampling distance (GSD) of 0.5 meters. The dataset features 31,333 annotated instances of binary changes.}

\bird{S2Looking~\cite{S2Looking} comprises 5,000 pairs of bi-temporal images captured by optical satellites worldwide. Each image is 1024 × 1024 pixels, with a GSD ranging from 0.5 to 0.8 meters. The dataset contains annotations for over 65,920 instances of binary changes.}

\subsubsection{Change Detection Results}
\bird{In change detection experiments, LSKNets are defaulting to be built within the Changer~\cite{changer} framework due to its compelling performance and open-source availability.
We conduct a comprehensive comparison of our proposed models, LSKNet-T and LSKNet-S, against 17 recent high-performance models on LEVIR-CD and S2Looking datasets. The results given in Tab.~\ref{tab:change} justify that the proposed LSKNet-T and LSKNet-S models have very compelling performance, surpassing all other state-of-the-art methods across all datasets in all main metrics (F1 and IoU).}

\subsection{Ablation Study}
In this section, we report ablation study results on the DOTA-v1.0 test set. 
The choice of the DOTA-v1.0 dataset for the ablation study is motivated by 2 factors. Firstly, object detection is known to be a practical and challenging task, and the DOTA-v1.0 dataset provides a diverse and complex set of objects and scenes for evaluation. Secondly, the availability of numerous models allows for comprehensive comparisons, enabling a thorough assessment of the effectiveness of our proposed method.
In ablation studies, we adopt the 100-epoch backbone pretraining schedule for experimental efficiency (Tab.~\ref{tab:ablation_lk_decomp},~\ref{tab:ablation_key_comps},~\ref{tab:max_avg_pool},~\ref{tab:frames_comp},~\ref{tab:lk_sm_comp}).

\begin{table}[t]
\renewcommand\arraystretch{1.2} 
\footnotesize
\setlength{\tabcolsep}{1.5mm}
\centering\caption{ \textbf{The effects of the number of decomposed large kernels} on the inference FPS and mAP, given the theoretical receptive field is 29. Decomposing the large kernel into two depth-wise kernels achieves the best performance of speed and accuracy.} 
\label{tab:ablation_lk_decomp}
  \vspace{4pt}
\begin{tabular}{lccll} 
($k$, $d$) sequence                                                                         & RF  & Num. & FPS  &mAP~(\%) \\ 
\Xhline{1pt}
\begin{tabular}[c]{@{}l@{}}(29, 1)\end{tabular}                               & 29 & 1  & 18.6 & 80.66\\
\begin{tabular}[c]{@{}l@{}}(5, 1) $\longrightarrow$ (7, 4)\end{tabular}                & 29   & 2   & \textbf{20.5} & \textbf{80.91}\\
\begin{tabular}[c]{@{}l@{}}(3, 1) $\longrightarrow$ (5, 2) $\longrightarrow$ (7, 3)\end{tabular}  & 29   & 3   & 19.2 & 80.77\\
\end{tabular}
\end{table}

\begin{table}[t]
    \renewcommand\arraystretch{1.2} 
    \setlength{\tabcolsep}{0.4mm}
    \footnotesize
    \centering  
    \caption{\textbf{The effectiveness of the key design components} of the LSKNet when the large kernel is decomposed into a sequence of two depth-wise kernels. CS: channel selection; SS: spatial selection \textbf{(ours)}. The LSKNet achieves the best performance when using a reasonably large receptive field with spatial selection. 
    }
    \label{tab:ablation_key_comps} 
    \begin{tabular}{ccccc|cccl} 
    ($k_1$, $d_1$) & ($k_2$, $d_2$) & Flow & CS & SS & RF & FPS & mAP  \\ 
    \Xhline{1pt}
    
    (3, 1)   & (5, 2)  &  Series &  -  &  -  &  11  & 22.1 & 80.80 \\ 
    (5, 1)   & (7, 3)  &  Series &  -  &  -  &  23  & 21.7  & 80.94 \\
    (5, 1)   & (7, 4)  &  Series &  -  &  -  &  29  & 20.5  & 80.91 \\
    (7, 1)   & (9, 4)  &  Series &  -  &  -  &  39  &  21.3 & 80.84\\ 
    \hline
    (3, 1)   & (5, 1)  &  Parallel &  $\checkmark$  &  -  &  5  &  23.3 & 80.19 &(SKNet\cite{li2019selective})\\  
    (5, 1)   & (7, 3)  &  Series & $\checkmark$  &  -  &  23  &  19.6 & 80.57  &(LSKNet-CS)\\
    (5, 1)   & (7, 3)  &  Series &  $\checkmark$  & $\checkmark$  &   23 & 18.6 & 80.82  &(LSKNet-SCS)\\
    (5, 1)   & (7, 3)  &  Series &  -  & $\checkmark$  &   23 & 20.7 & \textbf{81.31}  &(LSKNet) \\ 
    \end{tabular} 
\end{table}

\textbf{Large Kernel Decomposition.}
Deciding on the number of kernels to decompose is a critical choice for the LSK module. 
We follow Eq.~\eqref{eqn:dr} to configure the decomposed kernels.
The results of the ablation study on the number of large kernel decompositions, when the theoretical receptive field is fixed at 29, are shown in Tab.~\ref{tab:ablation_lk_decomp}. 
It suggests that decomposing the large kernel into two depth-wise large kernels results in a good trade-off between speed and accuracy, achieving the best performance in terms of both FPS (frames per second) and mAP (mean average precision).

\textbf{Kernel Receptive Field Size.}
Based on our evaluations presented in Tab.~\ref{tab:ablation_lk_decomp}, we find that decomposing the large kernel into two depth-wise kernels in \textbf{series} is optimal. Furthermore, Tab.~\ref{tab:ablation_key_comps} shows that excessively small or large receptive fields can hinder the performance of the LSKNet, and a receptive field size of approximately 23 is determined to be the most effective.

\begin{table}[t]
\renewcommand\arraystretch{1.2} 
\footnotesize
\setlength{\tabcolsep}{16pt}
\centering
\caption{Ablation study on the effectiveness of the \textbf{maximum and average pooling in spatial selection} of our proposed LSK module. The best result is obtained when using both.}
\label{tab:max_avg_pool} 
\begin{tabular}{cc|c|c} 
\multicolumn{2}{c|}{Pooling} & \multirow{2}{*}{FPS} & \multirow{2}{*}{mAP~(\%)}  \\ 
Max. & Avg.                  &                      &                       \\ 
\Xhline{1pt}
$\checkmark$    &                       &          20.7            &      81.23                 \\
     & $\checkmark$                     &           20.7       &        81.12               \\
$\checkmark$ & $\checkmark$          &             20.7       &            \textbf{81.31}           \\
\end{tabular}
\end{table}

\begin{table*}[th]
\renewcommand\arraystretch{1.3} 
\centering
\caption{\textbf{Comparison on LSKNet-S and other (large kernel or dynamic/selective attention) backbones} in remote sensing object detection (on DOTA-v1.0), segmentation (on Vaihingen) and change detection (on LEVIR-CD). Our LSKNet achieves the best mAP under similar or less complexity budgets. 
}
\label{tab:lk_sm_comp}
  \vspace{4pt}
\footnotesize
\setlength{\tabcolsep}{1mm}
\begin{tabular}{c|lcc|ccc|ccc|cccc} 

&&& & \multicolumn{3}{c|}{\textbf{\underline{~~~DOTA-v1.0~~~}}} & \multicolumn{3}{c|}{\textbf{\underline{~~~Vaihingen~~~}}} & \multicolumn{4}{c }{\textbf{\underline{~~~~~~LEVIR-CD~~~~~~}}}  \\
\multirow{-2}{*}{Group} & \multirow{-2}{*}{\begin{tabular}[c]{@{}l@{}}~~Model\\ Backbone\end{tabular}} & \multirow{-2}{*}{\textbf{\#P}} & \multirow{-2}{*}{\textbf{Flops}} & \textbf{mAP} & \textbf{@50} & \textbf{@75} &  \textbf{~F1~} & \textbf{~OA~} & \textbf{mIoU} & \textbf{P.} &  \textbf{R.} & \textbf{F1} & \textbf{IoU}  \\ 

\Xhline{1pt}
Baseline & ResNet-18  &  11.2M   &   ~38.1G  & 50.54 &  79.27 & 55.33 & 90.15 & 92.62 & 82.47 &92.97&90.61&91.77&84.80\\
\hline
\multirow{6}{*}{\begin{tabular}[c]{@{}c@{}}Large \\ Kernel\end{tabular}}       
& ViTDet~\cite{li2022exploring}     & 86.6M  & 394.9G  & 45.60 & 74.41 & 49.39 &81.01&83.74&54.91&80.72&90.59&85.37&74.48 \\  
& ConvNeXt v2-N~\cite{Woo2023ConvNeXtVC} &  15.0M   &   ~51.2G  & 52.91 & 80.81 & 58.58 &89.13&92.15&81.17&93.12&89.73&91.39&84.15 \\
& Swin-T~\cite{Swin} &   28.3M   &   ~ 91.1G & 51.54 & 80.81  & 56.71 &90.74&93.01&83.40&93.04&90.25&91.63&84.55\\
& MSCAN-S~\cite{guo_segnext_2022}   &  13.1M     &   ~45.0G    & 52.52 & 81.12  & 57.92 &91.16&93.04&84.10&93.39&91.14&92.25&85.62 \\
& VAN-B1~\cite{guo_visual_2022}     & 13.4M  & ~52.7G  & 52.69 & 81.15   & 58.11&91.30&93.12&84.41&93.31&91.20&92.24&85.60 \\  
\hline
\multirow{4}{*}{\begin{tabular}[c]{@{}c@{}} Dynamic/ \\ Selective \\ Attention\end{tabular}} 
& ResNeSt-14~\cite{resnest}    &   ~8.6M  &   ~57.9G   &49.79  & 79.51   & 53.41 &90.31&92.84&82.72&92.47&90.38&91.41&84.18 \\  
& SCNet-18~\cite{liu_improving_2020}      &   14.0M   &  ~50.7G  &  49.91 &    79.69 & 53.55&90.50&92.97&83.04&92.03&91.27&91.65&84.58 \\
& DCN-Res50~\cite{dai2017deformable}      &    26.2M   &  121.2G    & 49.26 & 79.74 & 52.97 &90.93&93.07&83.72& 92.84&90.67&91.74&84.74 \\
& SKNet-26~\cite{li2019selective}    & 14.5M & ~58.5G & 51.53 &80.67& 56.51 &90.83&93.01&83.56&93.09&91.09&92.08&85.32\\  
  
\hline
\rowcolor[rgb]{0.9,0.9,0.9} \textbf{Ours} & $\star$ LSKNet-S      &    14.4M   &  ~54.4G   & \textbf{53.32} &  \textbf{81.48}  & \textbf{58.83} & \textbf{91.81} & \textbf{93.61} & \textbf{85.12} &\textbf{93.44}&\textbf{91.13}&\textbf{92.27}&\textbf{85.65} \\ 
\end{tabular}
\end{table*}

\begin{table}[t]
\renewcommand\arraystretch{1.2} 

\footnotesize
\setlength{\tabcolsep}{1.8mm}
\centering
\vspace{-6pt}
\caption{\textbf{Comparison of LSKNet-T and ResNet-18} as backbones with different detection frameworks on DOTA-v1.0. The lightweight LSKNet-T achieves significantly higher mAP in various frameworks than ResNet-18. }
\label{tab:frames_comp}
  \vspace{4pt}
\begin{tabular}{l|cc} 
Frameworks  & ResNet-18 &  $\star$ LSKNet-T \\ 
\Xhline{1pt}
ORCNN~\cite{xie_oriented_2021}    &  79.27  &  81.31~{\textbf{(+2.04)}} \\
RoI Trans.~\cite{ding_learning_2019} &  78.32  &   80.89~{\textbf{(+2.57)}}   \\
\begin{tabular}[c]{@{}l@{}}
S$^2$A-Net~\cite{han_align_2020} \\\end{tabular} &  76.82  &    80.15~{\textbf{(+3.33)}}  \\
R3Det~\cite{yang_r3det_nodate}       &   74.16   &  78.39~{\textbf{(+4.23)}} \\ 
\hline
\#P {\scriptsize(backbone only)}  &    ~11.2M    &  ~~4.3M \textbf{{(-62\%)} }   \\
FLOPs {\scriptsize(backbone only)}  &    38.1G   &    19.1G \textbf{{(-50\%) }} \\
\end{tabular}
\end{table}

\textbf{Comparison with SKNet and Different Attention Selection Types.} 
There are two key distinctions between SKNet and LSKNet. Firstly, the proposed selective mechanism relies on explicit feature flow through a \textbf{series} of large kernels via decomposition, which deviates from the approach taken by most existing attention-based methods. In contrast, SKNet employs \textbf{parallel} decomposition. Secondly, LSKNet adaptively aggregates information across large kernels in the spatial dimension, instead of the channel dimension utilized in SKNet or LSKNet-CS. This design is more intuitive and effective for remote sensing tasks, as channel-wise selection fails to capture the spatial variance of different targets across the image space. \bird{Additionally, we evaluate a variant of LSKNet that leverages both spatial and channel selection.} Our experiments in Tab.~\ref{tab:ablation_key_comps} suggest that, in detection tasks, spatial information plays a more critical role. \bird{However, the inclusion of both spatial and channel selection may introduce extra difficulty in model optimization, leading to a slight performance drop.} A comprehensive conceptual comparison of the module architectures of SKNet, LSKNet, LSKNet-CS (channel selection version) and \bird{LSKNet-SCS (spatial and channel selection version)} is presented in Supplementary Materials.

\begin{figure*}[t]
  \centering
  \includegraphics[width=0.93\linewidth]{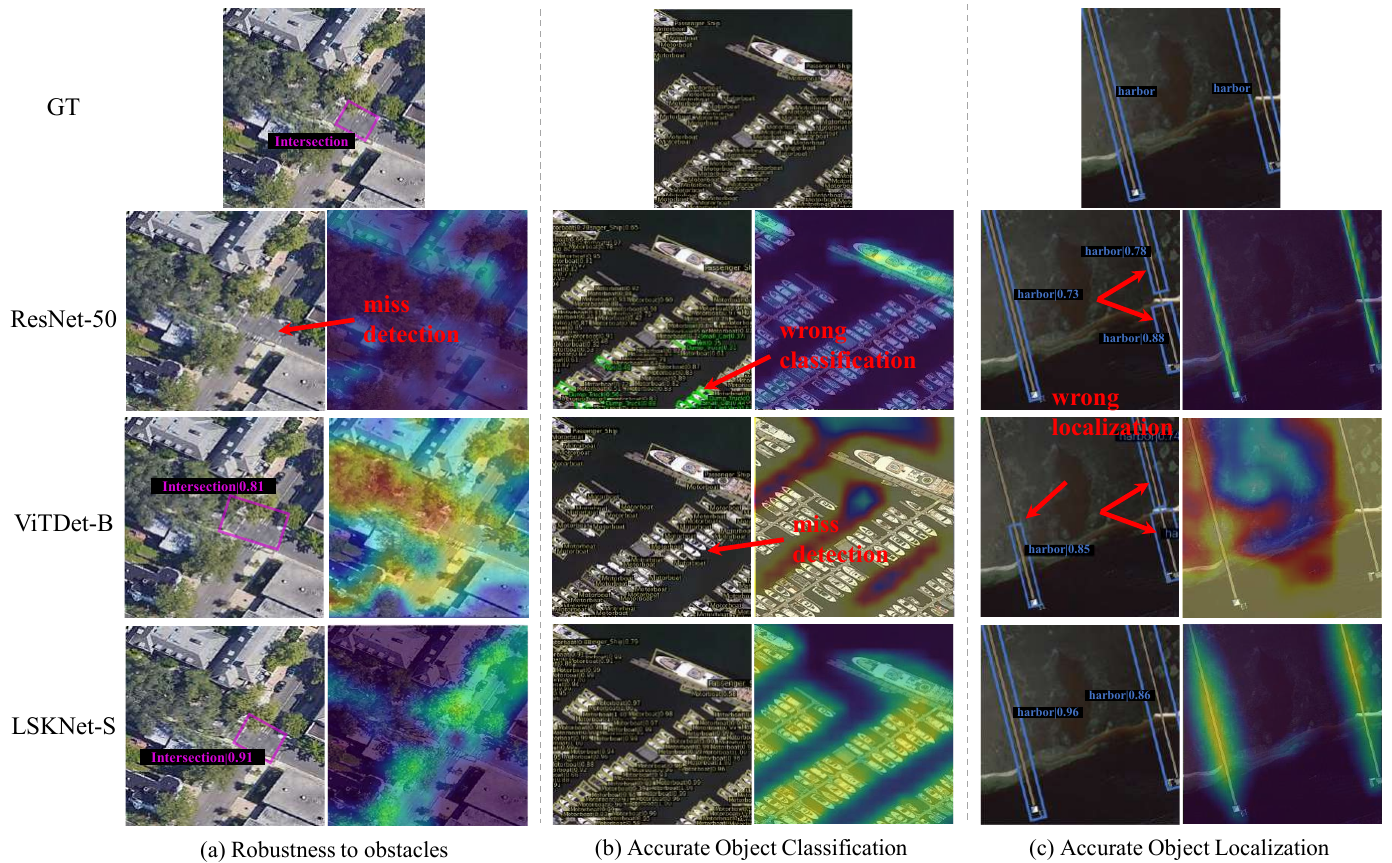} 
  \vspace{4pt}
  \caption{\textbf{Eigen-CAM visualization} of Oriented RCNN detection framework with ResNet-50, ViTDet and LSKNet-S. Our proposed LSKNet can model a reasonably long range of context information, leading to better performance in various hard cases.}
\label{fig:cam}
\end{figure*}

\textbf{Pooling Layers in Spatial Selection.}
We conduct experiments to determine the optimal pooling layers for spatial selection, as reported in Tab.~\ref{tab:max_avg_pool}. The results suggest that using both max and average pooling in the spatial selection component of our LSK module provides the best performance without sacrificing inference speed.

\textbf{Performance of LSKNet backbone under different detection frameworks.}
To validate the generality and effectiveness of our proposed LSKNet backbone, we evaluate its performance under various remote sensing detection frameworks, including two-stage frameworks O-RCNN~\cite{xie_oriented_2021} and RoI Transformer~\cite{ding_learning_2019} as well as one-stage frameworks S$^2$A-Net~\cite{han_align_2020} and R3Det~\cite{yang_r3det_nodate}.
The results in Tab.~\ref{tab:frames_comp} show that our proposed LSKNet-T backbone significantly improves detection performance compared to ResNet-18, while using only 38\% of its parameters and with 50\% fewer FLOPs. These findings underscore the lightweight yet powerful generality nature of the proposed LSKNet backbone.

\textbf{Comparison with Other Large Kernel/Selective Attention Backbones.}
We also compare our LSKNet with 9 popular or high-performance backbone models with large kernels or dynamic/selective attention. As shown in Tab.~\ref{tab:lk_sm_comp}, \bird{the ViTDet~\cite{li2022exploring}, which uses the vanilla ViT~\cite{dosovitskiy2020image} backbone, has the largest model size and computational complexity among the compared models and performs poorly across all tasks. Observations in Tab.~\ref{tab:dota10} show that it performs particularly poorly on objects with distinct fine-grained features (such as ball courts and Helicopter). It suggests that global contextual information is not as efficient or informative for remote sensing scenarios.} Under similar or fewer model sizes and complexity budgets, our LSKNet outperforms all other models in remote sensing object detection (on DOTA-v1.0), segmentation (on Vaihingen) and change detection (on LEVIR-CD), \bird{highlighting its effectiveness in capturing and processing semantic features in remote sensing images.} 

\section{Analysis}
We perform analysis specifically focused on the object detection task due to the significance of instance-level information in understanding the overall behaviour of the model.

\textbf{Detection Results Visualization. }
Visualization examples of detection results and Eigen-CAM~\cite{eign_cam} are shown in Fig.~\ref{fig:cam}. LSKNet can capture a reasonable range of context information relevant to the detected targets, leading to better performance in various hard cases, which justifies our \textit{prior 1)}. \bird{In contrast, ResNet typically captures only a limited range of context information, while ViTDet captures a large range but coarse spatial information, making it challenging to model fine-grained details when objects are small and crowded. Both models exhibit limited performance in challenging scenarios.}

\begin{figure}[t]
  \centering
  \includegraphics[ width=0.95\columnwidth]{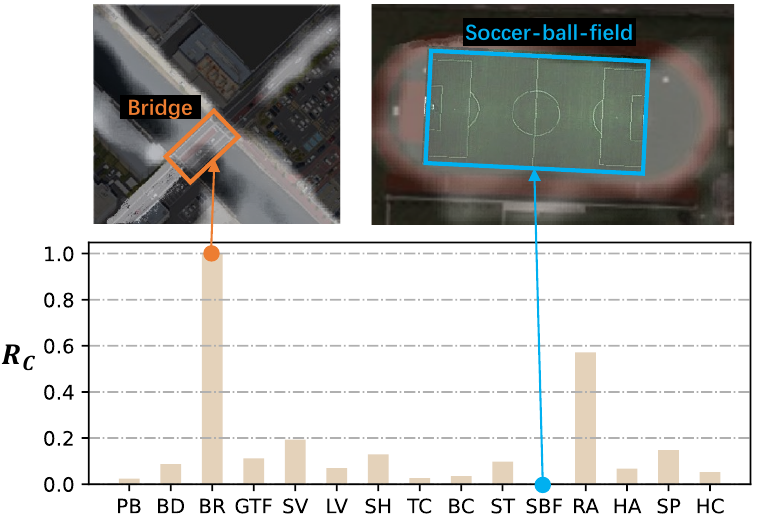}
  \vspace{4pt}
  \caption{Normalised \textbf{Ratio ${R_c}$ of Expected Selective RF Area and GT Bounding Box Area} for object categories in DOTA-v1.0. The relative range of context required for different object categories varies a lot. The visualized receptive field is obtained from Eq.~\eqref{eqn:7} (i.e., the spatial activation) of our well-trained LSKNet model.}
\label{fig:ratio}
\end{figure}

\textbf{Relative Context Range for Different Objects. }
To investigate the relative range of receptive field for each object category, we define $R_c$ as the \textit{Ratio of Expected Selective RF Area and GT Bounding Box Area } for category $c$:

\begin{equation}
   R_c = \frac{\sum_{i=1}^{I_c}{A_i / B_i}}{I_c} \text{,}
    \label{eqn:a_1}
\vspace{-4pt}
\end{equation}
\begin{equation}
    A_i = \sum_{d=1}^D{\sum_{n=1}^N{\lvert \widetilde{\mathbf{SA}}^d_n \cdot RF_n\rvert}},~ B_i = \sum_{j=1}^{J_i}{Area(\text{GT}_{j})} \text{,}
    \label{eqn:a_2}
\end{equation}

\noindent where $I_c$ is the number of images that contain the object category $c$ only. The $A_i$ is the sum of spatial selection activation in all LSK blocks for input image $i$, where $D$ is the number of blocks in an LSKNet, and $N$ is the number of decomposed large kernels in an LSK module. $B_i$ is the total pixel area of all $J_i$ annotated oriented object bounding boxes (GT). 
The normalized $R_c$ in Fig.~\ref{fig:ratio} represents the relative range of context required for different object categories for a better view. 

The results suggest that the Bridge category stands out as requiring a greater amount of additional contextual information compared to other categories, primarily due to its similarity in features with roads and the necessity of contextual clues to ascertain whether it is enveloped by water. 
\bird{Similarly, the roundabout category also has a relatively high $R_c$ of 0.57. Conversely, the Court categories have relatively low $R_c$ values, all lower than 0.1. They necessitate minimal contextual information due to their distinctive textural attributes, specifically the court boundary lines.}
It aligns with our knowledge and further supports \textit{prior 2)} that the relative range of contextual information required for different object categories varies greatly.

\begin{figure}[t]
  \centering
  \includegraphics[width=0.93\columnwidth]{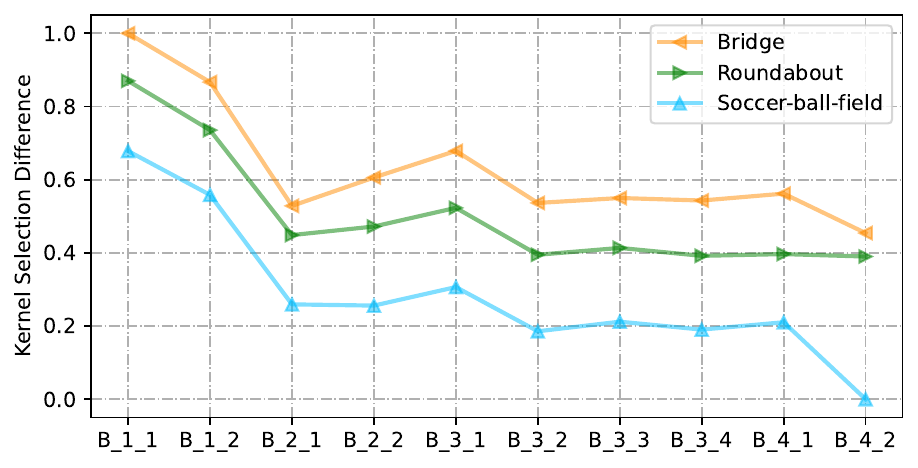}
  \vspace{4pt}
  \caption{Normalised \textbf{Kernel Selection Difference} in the LSKNet-T blocks for Bridge, Roundabout and Soccer-ball-field. B\_i\_j represents the j-th LSK block in stage i. A greater value is indicative of a dependence on a broader context.}
\label{fig:act_diff}
\end{figure}

\begin{figure*}[t]
\centering
\includegraphics[width=0.78\linewidth]{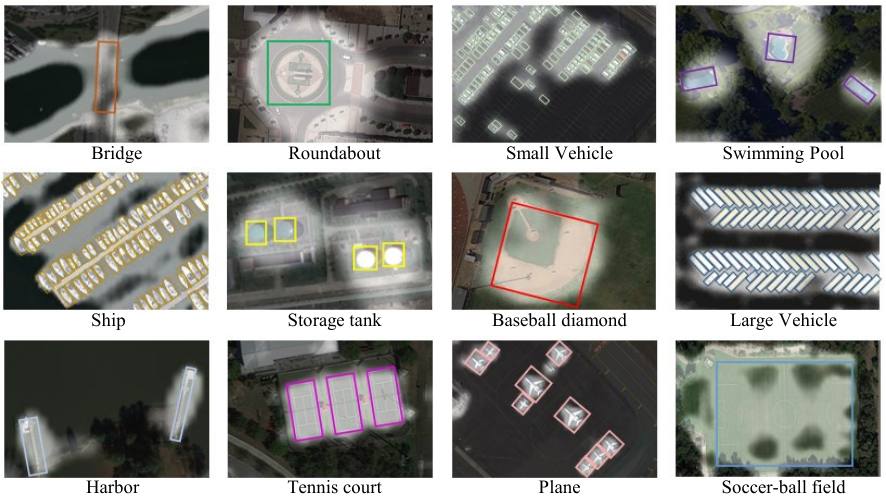}
  \vspace{4pt}
   \caption{Receptive field activation for more object categories in DOTA-v1.0, where the activation map is obtained from the Eq.~\eqref{eqn:7} (i.e., the spatial activation) of our well-trained LSKNet model.}
\label{fig:more_act} 
\end{figure*}

\textbf{Kernel Selection Behaviour. }
We further investigate the kernel selection behaviour in our LSKNet. For object category $c$, the \textit{Kernel Selection Difference} $\Delta A_c$ (i.e., larger kernel selection - smaller kernel selection)
of an LSKNet-T block is defined as:
\begin{equation}
    \Delta A_c = \lvert\widetilde{\mathbf{SA}}_{larger} - \widetilde{\mathbf{SA}}_{smaller}\rvert  \text{ .}
    \label{eqn:a_4}
\end{equation}
We demonstrate the normalized $\Delta A_c$ over all images for three typical categories: Bridge, Roundabout and Soccer-ball-field and for each LSKNet-T block in Fig.~\ref{fig:act_diff}. \bird{As expected, the $\Delta A_c$ of all blocks for Bridge is higher than that of Roundabout by about 30\% on average, and Roundabout is higher than Soccer-ball-field by about 70\%. }
This aligns with the common sense that Soccer-ball-field indeed does not require a large amount of context, since its own texture characteristics are already sufficiently distinct and discriminatory.

We also surprisingly discover another selection pattern of LSKNet across network depth: LSKNet usually utilizes larger kernels in its shallow layers and smaller kernels in higher levels. \bird{The average $\Delta A_c$ for the first layer blocks is 0.78, while for the second and third blocks it is 0.40, and for the last layer blocks it is only 0.33.} This indicates that networks tend to quickly focus on capturing information from large receptive fields in low-level layers so that higher-level semantics can contain sufficient receptive fields for better discrimination. 
 
\textbf{Spatial Activation Visualisations. }
Spatial activation map examples for more object categories in DOTA-v1.0 are shown in Fig.~\ref{fig:more_act}, where the activation map is obtained from Eq.~\eqref{eqn:7} (i.e., the spatial activation) of our well-trained LSKNet model. The object categories are arranged in decreasing order from top left to bottom right based on the \textit{Ratio of Expected Selective RF Area and GT Bounding Box Area } as illustrated in Fig.~\ref{fig:ratio}. The spatial activation visualization results also demonstrate that the model’s behaviour aligns with our proposed two priors and the above analysis, which in turn verifies the effectiveness of the proposed mechanism.

\section{Conclusion}
In this paper, we propose the lightweight Large Selective Kernel Network (LSKNet) as a novel approach for tackling downstream tasks in remote sensing images, such as scene classification, object detection, and semantic segmentation. LSKNet is specifically designed to leverage the inherent characteristics of remote sensing images: the need for a wider and adaptable contextual understanding. By adapting a large spatial receptive field, LSKNet can effectively capture and model diverse contextual nuances exhibited by different object types in remote sensing images. 
Extensive experiments demonstrate that our proposed lightweight model achieves state-of-the-art performance on competitive remote sensing benchmarks. The comprehensive analysis conducted throughout the paper validates the effectiveness and significance of our proposed lightweight model.\\

\section*{Acknowledgement}

This research was supported by the National Key Research and Development Program of China (No. 2021YFB3100800), Young Scientists Fund of the National Natural Science Foundation of China (Grant NO. 62206134, 62361166670, 62276145, 62176130, 62225604, 62301261),
the Fundamental Research Funds for the Central Universities 
(Nankai University, 070-63233084, 070-63233089), the Tianjin Key Lab of VCIP.
Computation is supported by the Supercomputing Center of Nankai University, and the China Postdoctoral Science Foundation (NO. 2021M701727).

\section*{Data Availability Statement}
\textbf{Data publicly available in a repository:}

The Imagenet dataset is available at \url{https://www.image-net.org/} 

The UCM dataset is available at \url{http://weegee.vision.ucmerced.edu/datasets/landuse.html}

The AID dataset is available at \url{https://captain-whu.github.io/AID/}

The NWPU dataset is available at \url{https://www.tensorflow.org/datasets/catalog/resisc45}

The MillionAID dataset is available at \url{https://captain-whu.github.io/DiRS/}

The DOTA dataset is available at \url{https://captain-whu.github.io/DOTA/dataset.html}

The FAIR1M-v1.0 dataset is available at \url{https://www.gaofen-challenge.com/benchmark}

SAR-Aircraft dataset is available at: \url{https://radars.ac.cn/web/data/getData?dataType=SARDataset\_en}

The Potsdam and Vaihingen datasets are available at \url{https://www.isprs.org/education/benchmarks/UrbanSemLab/default.aspx}

The LoveDA dataset is available at \url{https://codalab.lisn.upsaclay.fr/competitions/421}

The UAVid dataset is available at \url{https://uavid.nl/}

The GID dataset is available at \url{https://x-ytong.github.io/project/GID.html}

The LEVIR-CD dataset is available at \url{https://justchenhao.github.io/LEVIR/}

The S2Looking dataset is available at \url{https://github.com/S2Looking/Dataset}

\backmatter

\bibliography{sn-bibliography}

\end{document}